%% file: main.tex
\documentclass[runningheads]{llncs}

\usepackage{eccv}

\usepackage{eccvabbrv}

\usepackage{graphicx}
\usepackage{booktabs}
\usepackage{xspace}
\usepackage[dvipsnames]{xcolor}
\usepackage{color, colortbl}

\usepackage{amsmath}
\usepackage{makecell}
\usepackage{multirow}
\usepackage{float}
\usepackage{amssymb}
\usepackage{utfsym}
\usepackage{diagbox}
\usepackage{dsfont}
\usepackage[numbers,sort&compress]{natbib}

\newcommand{\hydra}{ZTRS\xspace}
\newcommand{\gc}{\cellcolor{gray!15}}

\makeatletter
\newcommand\blfootnote[1]{%
  \begingroup
  \renewcommand\@makefnmark{}%
  \footnotetext{#1}%
  \endgroup
}
\makeatother

\usepackage[accsupp]{axessibility}  

\usepackage{hyperref}

\usepackage{orcidlink}

\begin{document}

\title{Zero-Human Demonstration End-to-end Autonomous Driving with Trajectory Scorer} 

\titlerunning{Zero-Human Demonstration End-to-end Autonomous Driving}



\author{
Zhenxin Li\inst{1,2*}\orcidlink{0009-0004-7430-9593} \and
Nadine Chang\inst{3}\orcidlink{0000-0003-4765-8478} \and
Wenhao Yao\inst{1,2}\orcidlink{0009-0000-5337-1663} \and
Xinglong Sun\inst{3}\orcidlink{0009-0003-8161-6922} \and
Zi Wang\inst{3}\orcidlink{0009-0002-7129-1928} \and
Maying Shen\inst{3}\orcidlink{0009-0000-9416-680X} \and
Jingde Chen\inst{3}\orcidlink{0009-0007-4408-8373} \and
Jingyu Song\inst{4}\orcidlink{0000-0001-8458-4303} \and
Kailin Li\inst{5}\orcidlink{0009-0004-0358-5240} \and \\
Zuxuan Wu\inst{1,2\dagger}\orcidlink{0000-0002-8689-5807} \and
Shiyi Lan\inst{3}\orcidlink{0000-0003-1183-6690} \and
Jose M. Alvarez\inst{3}\orcidlink{0000-0002-7535-6322}
}

\authorrunning{Z. Li et al.}

\institute{
Institute of Trustworthy Embodied AI, Fudan University \and
Shanghai Key Laboratory of Multimodal Embodied AI \and
NVIDIA \and
University of Michigan \and
East China Normal University
}

\maketitle

\blfootnote{$^{*}$Work done during an internship at NVIDIA.
$^{\dagger}$Corresponding author.
}

\input{sec/0_abstract}

\input{sec/1_intro}

\input{sec/2_related}
\input{sec/3_method}
\input{sec/4_exp}

\input{sec/5_conclusion}

\clearpage
\appendix
\input{supp_content}

%
%
\bibliographystyle{splncs04}
\bibliography{main}

\end{document}

%% file: sec/0_abstract.tex
\begin{abstract}

Human demonstrations are widely considered the cornerstone of end-to-end (E2E) autonomous driving despite human demonstration's scarcity for long-tail and safety-critical scenarios. Nonetheless, current E2E autonomous driving (AD) training paradigms continue to rely on human demonstrations. Imitation learning (IL) requires human demonstrations for training, whereas reinforcement learning (RL) has emerged as a promising alternative to reduce this dependency. However, most existing RL methods for E2E AD still rely implicitly on human demonstrations. A pure rewards-based RL method can overcome the need for human demonstrations, but general RL policy gradient methods suffer from the cold-start problem. In this paper, we propose ZTRS (\textbf{Z}ero-human demonstration end-to-end autonomous driving with \textbf{TR}ajectory \textbf{S}corer) — a complete RL-based E2E planning paradigm trained solely on real-world images and rule-based rewards, entirely without human demonstration. Through our proposed \textbf{Exhaustive Policy Optimization (EPO)}, a policy gradient variant tailored for enumerable trajectory actions and dense supervision, ZTRS enables the model to generalize better to long-tail driving scenarios. We demonstrate this generalization through our SOTA performance against IL approaches on both long-tail Navhard and closed-loop HUGSIM datasets. Project page: \url{https://zhenxinli.net/ZTRS/}.

\keywords{End-to-end Autonomous Driving \and Zero-Human Demonstration \and Exhaustive Policy Optimization}

\end{abstract}

%% file: sec/1_intro.tex
\section{Introduction}

\input{figs/paradigms}

Recently, end-to-end (E2E) planners~\cite{chitta2022transfuser, hu2023planning, jiang2023vad, chen2024vadv2, li2024hydra}, which take in multi-sensor data and output trajectories directly, are the standard approach to autonomous driving (AD). Specifically, two predominant training paradigms have emerged: Imitation Learning (IL) and Reinforcement Learning (RL). However, both E2E AD paradigms depend on human demonstrations, which are notably scarce in AD and especially so for safety-critical, long-tail scenarios. On one hand, IL relies exclusively on human demonstrations, leading to a number of additional critical issues: covariate shift~\cite{chang2021mitigating} and causal confusion~\cite{wen2020fighting}. On the other hand, RL methods attempt to mitigate IL's issues by training with reward signals--whether by finetuning a pre-trained IL model~\cite{gao2025rad, zou2025diffusiondrivev2, zhou2025autovla, noguchi2025offline, li2025finetuning, li2025recogdrive} or incorporating demonstrations into the reward signal, such as measuring L2 distances between planned and expert trajectories~\cite{gao2025rad, noguchi2025offline, li2025finetuning}.
As a result, bias towards human demonstrations persists and restricts the model’s ability to explore beyond the support of expert demonstrations. Ultimately, we encounter the main AD critical challenge in both IL and RL: inability to address safety-critical, long-tail cases where we do not have access to human demonstrations. Thus, in this work, we focus on rewards-based RL to overcome the reliance on human demonstrations. 

Generally in RL, current policy gradient methods can be used off-the-self to train a pure rewards-based E2E RL planner without human demonstrations. However, employing common policy gradient methods--such as PPO~\cite{schulman2017proximal, huang2022cleanrl}, REINFORCE~\cite{sutton1999policy}, or GRPO~\cite{shao2024deepseekmath, li2025finetuning}--to generate continuous actions often leads to the common RL cold-start problem: without structured guidance, a random agent struggles to find high-reward trajectories in a large, continuous space~\cite{nair2018overcoming}. Motivated by overreliance on human demonstration and cold-start limitation, we propose a fully from scratch RL-based \textbf{Z}ero-human demonstration end-to-end autonomous driving with \textbf{TR}ajectory \textbf{S}corer (ZTRS) to achieve: (1) a broad exploration landscape to cover long-tail scenarios and (2) robust and stable optimization.

As shown in ~\cref{fig:teaser}, ZTRS is a neural planner that selects actions from a trajectory vocabulary. The comprehensive nature of this vocabulary that covers nearly all driving actions has proven to improve trajectory generalization~\cite{philion2020lift, phan2020covernet, chen2024vadv2, li2024hydra}.
Instead of relying on stochastic exploration, the vocabulary is constructed using cluster anchors from offline datasets~\cite{philion2020lift}. 
Crucially, we note that we are not training with an explicit human demonstration for each sample, and the reconstructed vocabulary is a constant model input. Leveraging this vocabulary, we propose Exhaustive Policy Optimization (EPO) that trains the planner from scratch by performing an exhaustive evaluation across the entire vocabulary. This dense supervision approach ensures a higher probability of capturing sparse but high-reward trajectory patterns in large spaces, effectively resolving the cold-start problem. To keep the exhaustive evaluation tractable, we use rule-based metrics~\cite{dauner2024navsim, li2025hydra, cao2025pseudo} as reward signals, which are both efficient and well-suited to E2E AD setting. 
Furthermore, we show that our dense supervision framework facilitates stable reward-driven learning. We also provide a theoretical analysis on EPO's viability to optimize a dense, full action space.

Critically, EPO demonstrates the following benefit: its ability to generalize to long-tail cases. Since EPO enumerates through nearly all driving actions, it allows ZTRS to learn trajectories often unavailable as expert demonstrations--rare trajectories critical for navigating challenging long-tail cases. This enables ZTRS to better generalize to long-tail test cases and achieve state-of-the-art (SOTA) performance on open-loop safety-critical Navhard. Importantly, ZTRS also achieves SOTA in closed-loop HUGSIM, demonstrating that ZTRS is not merely ``hacking'' open-loop rewards.
Finally, while gaining long-tail generalizability, ZTRS also demonstrates that it can maintain SOTA performance on nominal driving. Our main contributions are as follows:

\begin{enumerate}
\item We introduce \textbf{\hydra}, a fully RL-based neural planner for E2E AD that operates without depending on explicit human demonstration. This work demonstrates the feasibility to plan purely from rule-based rewards.

\item We propose \textbf{Exhaustive Policy Optimization (EPO)}, an offline reinforcement learning method tailored for enumerable trajectory actions, offering dense supervision and stable convergence.

\item We demonstrate ZTRS's safety-critical, long-tail generalization by achieving SOTA on open-loop Navhard and closed-loop HUGSIM. We further illustrate how EPO resolves cold-start problem as compared to other RL policy methods. Lastly, we show that we are nonetheless able to maintain SOTA performance on par with IL methods for generic open-loop planning, Navtest.

\end{enumerate}

%% file: figs/paradigms.tex
\begin{figure*}[!t]
    \centering
    \includegraphics[width=\linewidth]{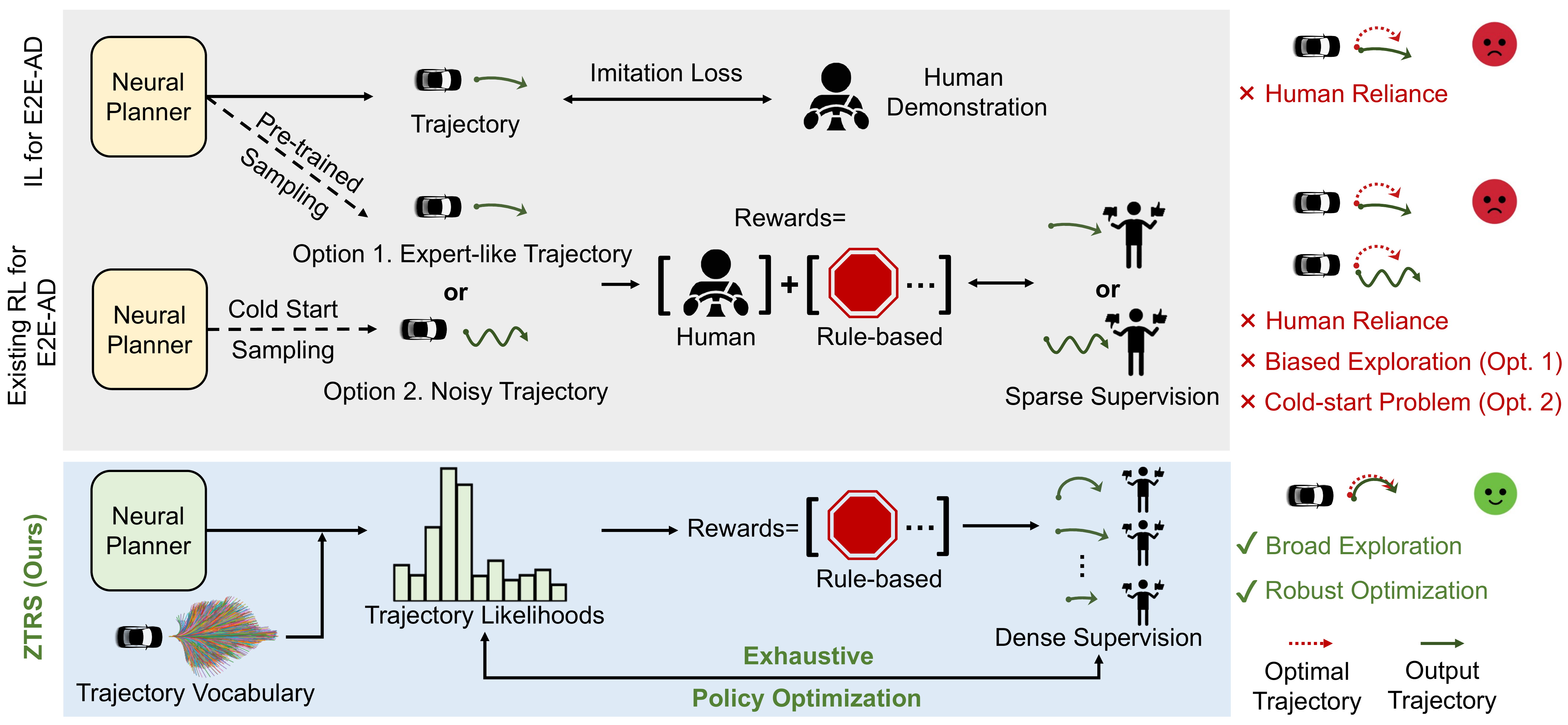}

    \caption{\textbf{Paradigms for Training E2E Neural Planners.} While current IL and RL rely on human-biased supervision, ZTRS is the first paradigm using zero-human demonstrations for training E2E neural planners. We leverage a dense trajectory vocabulary and Exhaustive Policy Optimization (EPO) to provide dense reward-based supervision across all potential trajectories. EPO effectively resolves the exploration constraints of IL-pretrained models and the cold-start problem of traditional RL, enabling ZTRS to generalize to rare, long-tail scenarios that are scarce in human datasets.}
    \vspace{-0.25in}

    \label{fig:teaser}
\end{figure*}

%% file: sec/2_related.tex
\section{Related Work}
\subsection{Imitation Learning for E2E-AD}

Given an offline expert dataset, Imitation Learning (IL) trains a policy to mimic expert behavior. In the end-to-end autonomous driving literature~\citep{chen2024end}, early IL-based approaches~\citep{codevilla2018end} are trained and tested in the closed-loop driving simulator CARLA~\citep{dosovitskiy2017carla}. Subsequent works improve the closed-loop driving performance with modern neural architectures (e.g. Transformers~\citep{vaswani2017attention})~\citep{chitta2022transfuser}, intermediate representations~\citep{hu2022model, Renz2022CORL, shao2023safety, jia2023think}, and policy distillation~\citep{chen2020learning, zhang2021end, wu2022trajectory, jia2023driveadapter}. The introduction of UniAD~\citep{hu2023planning} highlighted the strength of IL on real-world sensor data, whose complexity and diversity greatly exceed synthetic data produced by simulators. Building on this foundation, numerous efforts further focus on efficiency~\citep{jiang2023vad, liao2025diffusiondrive}, multi-modal behaviors~\citep{chen2024vadv2, liao2025diffusiondrive}, vision-language understanding~\citep{wang2024he, li2025recogdrive}, and safety constraints~\citep{li2024hydra}.

\vspace{-0.1in}
\subsection{Reinforcement Learning for AD}
Unlike Imitation Learning, Reinforcement Learning (RL)~\citep{sutton1998reinforcement} trains agents to maximize rewards by interacting with the environment, and autonomous driving RL methods generally fall into two categories: symbolic-input methods and sensor-based methods, both relying on simulators. Symbolic-input methods~\citep{toromanoff2020end, zhang2021end, li2024think2drive, booher2024cimrl, cusumano2025robust, jaeger2025carl} use low-dimensional abstractions (e.g. 3D bounding boxes, maps, traffic signals) and have shown strong results on CARLA~\citep{dosovitskiy2017carla}, nuPlan~\citep{nuplan}, and Waymax~\citep{gulino2023waymax}.
GigaFlow~\citep{cusumano2025robust} and CaRL~\citep{jaeger2025carl} both demonstrated that large-scale RL could train robust policies from scratch. On the other hand, sensor-based approaches operate on raw inputs like images. Early attempts~\citep{kendall2019learning} explored real-world training, but faced safety and efficiency challenges, while subsequent works~\citep{nehme2023safe, delavari2025comprehensive, yang2025raw2drive} relied on simulated sensor data in CARLA. Despite success in simulators, such approaches could face sim-to-real gaps. Recently, RAD~\citep{gao2025rad} fine-tuned a policy in a 3DGS simulator with high-dimensional real-world images, but still depended on human demonstrations for pre-training and reward computation. 
Similarly, several other approaches avoid the cold-start problem by fine-tuning an IL-pretrained diffusion policy~\citep{li2025recogdrive, li2025finetuning}.
In contrast, our approach learns trajectory planning with reward signals from scratch, while operating on high-dimensional real-world sensor data.

\vspace{-0.1in}
\subsection{End-to-end Trajectory Scorers}
End-to-end Trajectory Scorers introduce a predefined vocabulary containing multiple trajectories, and scores each trajectory to select the most appropriate candidate as the output. 
Early works~\citep{philion2020lift, phan2020covernet, chen2024vadv2} score the trajectory candidates through classification based on their distance towards the ground-truth human trajectory.
Beyond relying on a single human demonstration, the Hydra-MDP series~\citep{li2024hydra, li2025hydra} proposed multi-target hydra-distillation to score trajectories with multiple rule-based metrics, leading to more robust planning capability. SafeFusion~\citep{wang2025enhancing} synthesizes collision-related scenarios for training a robust planning model and eases the reliance on imitation learning. Other works further introduce multiple approaches to reach more precise and comprehensive trajectory scoring, such as test-time training~\citep{sima2025centaur}, iterative refinement~\citep{yao2025drivesuprim}, and diffusion-based trajectory generation~\citep{li2025hydranext, li2025generalized}.
Despite these advancements, these scorers still rely heavily on the imitation of human trajectories. In contrast, our approach fully discards expert demonstrations and achieves strong end-to-end planning performance via Reinforcement Learning.

%% file: sec/3_method.tex
\section{Methodology}

\input{figs/arch}

In this section, we elaborate on the framework and the Exhaustive Policy Optimization technique used in \hydra.

\subsection{Preliminary: Trajectory Scorers}
\label{sec:overall}
As shown in Fig.~\ref{fig:arch}, \hydra is a trajectory scorer~\citep{li2024hydra, li2025hydra, li2025hydranext, wang2025enhancing, sima2025centaur, yao2025drivesuprim, li2025generalized}, whose functionality is to score a discrete vocabulary of trajectories $\mathcal{A}=\{a_i\}_{i=1}^n$ instead of regressing to a continuous trajectory.
A dense, pre-collected vocabulary encompasses numerous trajectory candidates with natural driving patterns. In this formulation, the goal becomes selecting the most likely trajectory.

\hydra consists of five modules: an image backbone, a trajectory tokenizer, a Transformer Decoder, a policy head, and several scoring heads. The policy head produces likelihoods for taking each action in $\mathcal{A}$, while the scoring heads predict the scores of rule-based driving metrics.
For the choice of the metrics, we follow the practice of prior trajectory scorers (e.g. Hydra-MDP~\cite{li2024hydra} and GTRS~\cite{li2025generalized}) to use the Extended Predictive Driver Model Score (EPDMS, $\mathcal{E}$)~\citep{dauner2024navsim, li2025hydra, cao2025pseudo}. EPDMS comprehensively evaluates multiple aspects of driving (e.g., safety, progress, and rule compliance) and can be efficiently computed for the trajectories in $\mathcal{A}$. Given a state $s$ sampled from an offline dataset $\mathcal{D}$, where $s$ contains sensor data and ego-vehicle status, the forward process involves three steps:\begin{itemize}

    \item The image backbone extracts $L$ image tokens $\{x^i_{img}\}_{i=1}^L$ from a frontal-view image, while the trajectory tokenizer encodes trajectory candidates into queries $\{x_{traj}^i\}_{i=1}^n$. 
    \item In the Transformer Decoder, the trajectory queries attend to image tokens.
    \item The policy head maps the attended trajectory queries $\{x_{traj}^i\}_{i=1}^n$ to likelihoods $\pi(\cdot|s)$, and $m$ scoring heads map them to $m$ rule-based scores $\{\mathcal{S}_i(\cdot|s)\}_{i=1}^m$.

\end{itemize}
During training, the scoring heads are trained with binary classification losses against $\mathcal{E}(s,\cdot)$, while the policy head is trained with Exhaustive Policy Optimization (See Sec.~\ref{sec:solution}). At inference, the final trajectory $a\in\mathcal{A}$ is chosen using a weighted average of the likelihoods $\pi(\cdot|s)$ and predicted metric scores $\{\mathcal{S}_i(\cdot|s)\}_{i=1}^m$, following Hydra-MDP~\cite{li2024hydra, li2025hydra}.

\subsection{Revisiting the Policy Gradient Theorem}
\label{sec:prelim-pg}
To optimize a trajectory scorer with only rewards, we start with a simplified one-step policy optimization problem where the action space is a finite discrete set $\mathcal{A}$, 
and $\pi$ is a policy parameterized by $\theta$.
In the online RL setting~\citep{sutton1998reinforcement}, the action $a$ is sampled with $\pi(a|s)$, where $s$ is the current state. 
In our offline RL setting~\citep{levine2020offline}, the state $s$ is sampled from an offline dataset $\mathcal{D}$. 
Following the notation of GAE~\cite{schulman2015high}, the policy gradient $g$ is defined as
\begin{equation}
g := \mathbb{E} \left[ \Psi(s, a) \nabla_\theta \log \pi_\theta(a \mid s) \right], \label{eq:logll}
\end{equation}
where the advantage function $\Psi(s, a)$ can represent many quantities, such as the cumulative return or the state-action value function.
The gradient can be equivalently written as
\begin{align}
g 
&= \mathbb{E} \left[ \Psi(s, a) \nabla_\theta \log \pi_\theta(a \mid s) \right] \\
&= \sum_{a'\in \mathcal{A}} \Psi(s, a')\pi_\theta(a' \mid s)\,  \nabla_\theta \log \pi_\theta(a' \mid s) \\
&= \sum_{a'\in \mathcal{A}} \Psi(s, a')\pi_\theta(a' \mid s)\,  \frac{\nabla_\theta \pi_\theta(a' \mid s)}{\pi_\theta(a' \mid s)} \\
&= \sum_{a'\in \mathcal{A}} \Psi(s, a') \nabla_\theta \pi_\theta(a' \mid s). \label{eq:policy-gradient}
\end{align}
This formulation matches the classical Policy Gradient Theorem~\citep{sutton1999policy}. Notably, the summation over all actions in $\mathcal{A}$ suggests that if the advantage function $\Psi$ can be computed for each action at state $s$, policy optimization can be carried out directly on action likelihoods rather than log-likelihoods, as shown in Eq.~\ref{eq:policy-gradient}.

\subsection{Exhaustive Policy Optimization}
\label{sec:solution}
When the action space $\mathcal{A}$ is a set of trajectories covering almost all driving possibilities, policy optimization can be formulated as maximizing the objective for an offline dataset $\mathcal{D}$:
\begin{equation}
    \mathbb{E}_{s \sim \mathcal{D},\, a \sim \mathcal{A}} \left[ \Psi(s,a) \right].
    \label{eq:obj}
\end{equation}
This is consistent with the one-step policy optimization problem in Sec.~\ref{sec:prelim-pg} if $\Psi$ is derived from open-loop reward signals, which do not require additional environment interaction. This objective can be optimized either in the log-likelihood form on a sampled action (Eq.~\ref{eq:logll}) or in the likelihood form on the entire action space (Eq.~\ref{eq:policy-gradient}).
The latter formulation, \begin{equation}
g := \sum_{\substack{a' \in \mathcal{A} \\ s \sim \mathcal{D}}} \Psi(s, a') \, \nabla_\theta \pi_\theta(a' \mid s) \label{eq:epo}
\end{equation}
provides much denser supervision for the policy as it explicitly enumerates all possibilities. Eq.~\ref{eq:epo} also defines our proposed \textit{Exhaustive Policy Optimization (EPO)}, a variant of policy gradient tailored for offline data and enumerable actions. Specifically, EPO optimizes the policy by exhaustively considering every possible action $a\in\mathcal{A}$ and its respective advantage $\Psi(s,a)$. 
This formulation naturally enables broad exploration when the action space is $\mathcal{A}$ is dense. Meanwhile, EPO skips the sampling procedure, removing the need for IL pre-training and cold-start sampling. This dense, full-action-space optimization process is generally impractical in online RL but allows our framework to fully leverage off-policy trajectories for stable and exploratory learning. An overview of the pipeline is shown in Fig.~\ref{fig:arch}.

To compute the advantage function $\Psi$ without relying on human demonstrations, we again adopt the EPDMS metric $\mathcal{E}$ for its efficient and comprehensive evaluation of trajectories. 
Our empirical findings suggest that using the score $\mathcal{E}(s,\cdot)$ as $\Psi(s,\cdot)$ enables better long-tail, closed-loop generalization than baselines trained with human demonstrations.
Furthermore, the scores $\mathcal{E}(s,\cdot)$ can also be reused for each state $s\in\mathcal{D}$ as long as the action space is fixed throughout the training process, which greatly improves training efficiency.
To further enforce temporal consistency, we subtract a correction term $b(s_t, a_t, a_{t-1}) = \lambda\mathds{1}\left[\text{EC}(a_{t-1}, a_t)\right],$ where $\lambda$ is a constant, $a_{t-1}=\underset{a}{\text{argmax}}\space \pi(a|s_{t-1})$, and EC indicates the violation of Extended Comfort (EC) thresholds ~\citep{li2025hydra}: \begin{equation}
    \Psi(s_t,a_t)=\mathcal{E}(s_t,a_t)-b(s_t,a_t,a_{t-1}).
\end{equation}
Note that this formulation penalizes inconsistent predictions more strongly than the one used in~\citep{li2025hydra, cao2025pseudo}. Finally, $\Psi$ is normalized to zero mean and unit variance following~\citep{huang2022cleanrl, shao2024deepseekmath}.

%% file: figs/arch.tex
\begin{figure*}[tp]
    \centering
    \includegraphics[width=\linewidth]{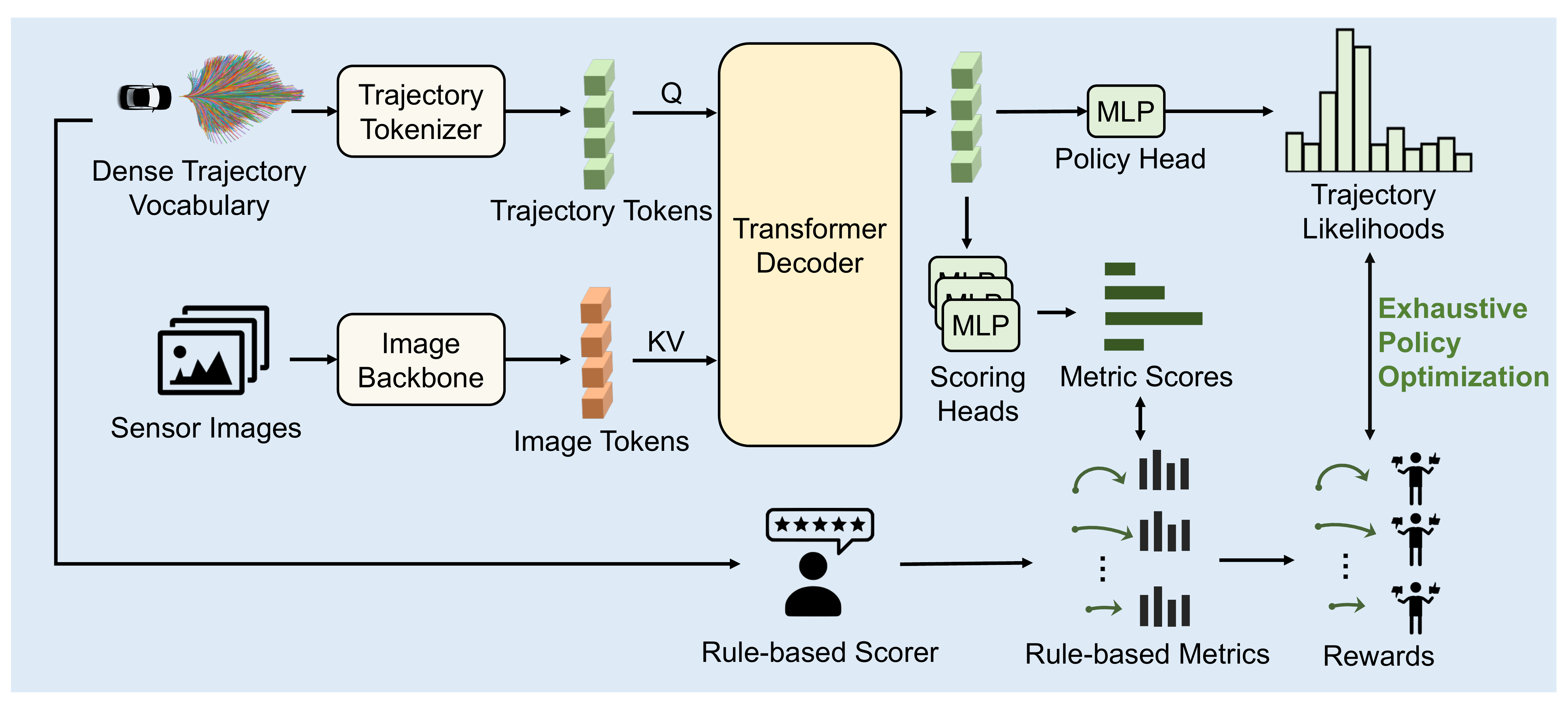}

    \caption{\textbf{The Overall Framework of \hydra.} Given offline sensor images and a fixed set of trajectories, \hydra first tokenizes these two modalities. In a Transformer Decoder, the trajectory tokens attend to image tokens to acquire the context. Finally, scoring heads and a policy head map the trajectory tokens to metric scores and action likelihoods. All components are trained in an end-to-end manner.}

    \label{fig:arch}
\end{figure*}

%% file: sec/4_exp.tex
\section{Experiments}

In this section, we quantitatively show that ZTRS can generalize to long-tail and safety-critical scenarios. Through our ablation studies, we show the necessity of our design choices. Finally, we qualitatively illustrate what type of long-tail and safety-critical cases ZTRS addresses.

\subsection{Datasets and metrics}
\label{sec:metrics}
\noindent\textbf{Datasets.}
We conduct experiments on three benchmarks covering open-loop, closed-loop, and long-tail testing: Navhard~\citep{cao2025pseudo},  HUGSIM~\citep{zhou2024hugsim}, and Navtest~\citep{dauner2024navsim}.

Our first dataset is Navhard, its long-tail and safety-critical scenarios makes it a suitable dataset to evaluation long-tail generalization. Navhard~\citep{cao2025pseudo} uses pseudo-simulation on the long-tail challenging scenarios in NAVSIM, resulting in the total evaluation to a two-stage paradigm: 1) first stage follows the original NAVSIM evaluation, 2) second stage adopts 3D Gaussian Splatting (3DGS)~\citep{kerbl20233d, li2025mtgs} to synthesize subsequent driving scenarios, resulting in 244 initial scenarios and 4164 synthetic scenarios. 

HUGSIM~\citep{zhou2024hugsim} is a closed-loop driving benchmark that does not contain human demonstrations; thus an appropiate dataset to test non-human demonstration method. It features images synthesized with 3DGS and integrates multiple driving datasets, including KITTI-360~\citep{liao2022kitti}, Waymo~\citep{sun2020scalability}, nuScenes~\citep{caesar2020nuscenes}, and Pandaset~\citep{xiao2021pandaset}, into a collection of 345 driving scenarios. These scenarios are categorized by difficulty into four levels: easy, medium, hard, and extreme. Specifically, HUGSIM released 49 easy scenarios for regular driving, 126 medium scenarios with inserted vehicles, and 86 hard scenarios as well as 84 extreme scenarios with aggressive vehicles.

\noindent\textbf{Metrics.} Navhard and Navtest evaluate open-loop planning with EPDMS $\mathcal{E}(s,a)$, which is defined as an aggregation of multiple driving metrics:
\begin{equation}
\mathcal{E}(s,a) = \left(
    \displaystyle\prod_{m \in S_{\mathrm{pen}}} m(s,a)
\right)
\cdot
\left(
    \frac{\sum_{m \in S_{\mathrm{avg}}} w_m \, m(s,a)}
         {\sum_{m \in S_{\mathrm{avg}}} w_m}
\right)
\label{eq:epdms}
\end{equation}
where $s$ is the current state and $a$ is a 4-second trajectory. The penalty metric set $S_{\mathrm{pen}}$ is applied multiplicatively and includes No-at-fault Collisions (NC), Drivable Area Compliance (DAC), Driving Direction Compliance (DDC), and Traffic Light Compliance (TLC), while the weighted metric set $S_{\mathrm{avg}}$ contains Time-to-Collision (TTC), Ego Progress (EP), Lane Keeping (LK), and History Comfort (HC). The extended comfort (EC) from~\citep{li2025hydra} is also used as a weighted metric to promote temporally consistent driving. $w_m$ is the aggregation weight for metric $m$. Note that human filtering is used in~\citep{cao2025pseudo} for calculating $\mathcal{E}$ but not in our training. For HUGSIM, HD-Score is used for closed-loop evaluation. It aggregates Route Completion (RC) with the sub-metrics NC, DAC, TTC, HC across an episode of length $T$: 
\begin{equation}
\resizebox{0.9\columnwidth}{!}{%
$
\text{HD-Score} = RC \cdot \sum_{t=1}^T \left[ \left( \prod_{m \in \{NC, DAC\}} m(s_t, \tilde{a}_t) \right) \cdot \left( \frac{ \sum_{m \in \{TTC, HC\}} w_m\, m(s_t, \tilde{a}_t) }{ \sum_{m \in \{TTC, HC\}} w_m } \right) \right],
$
}
\label{eq:hdscore}
\end{equation}
where $\tilde{a}$ is the ego-vehicle acceleration and steering angle transformed from a trajectory. 

Navtest~\citep{dauner2024navsim} is the common open-loop evaluation dataset for NAVSIM. We evaluate on this dataset to showcase ZTRS' ability to handle more nominal driving. NAVSIM contains 103k and 12k diverse and challenging driving scenarios for model training (Navtrain) and evaluation (Navtest), and introduces simulation-based metrics to better review closed-loop planning capability through open-loop evaluation. During evaluation, the output trajectory is evaluated by a simulator to get rule-based simulation metric scores related to multiple driving aspects, such as traffic rule compliance, comfort, and progress.

\subsection{Implementation Details}
All our models are trained on the Navtrain split with 24 NVIDIA A100 GPUs, while the synthetic data from Navhard and HUGSIM are not used for training. Models are trained for 15 epochs with a total batch size of 528, using a learning rate and weight decay of $2\times10^{-4}$ and 0.0. The frontal view with center-cropped front-left and front-right views are concatenated as the input image, which is then resized to $512\times2048$. The hyperparameter $\lambda$ in the correction term $b$ is set to $0.2$. By default, the action space used in our method has 16384 trajectories, each spanning 4 seconds at 10Hz. These trajectories are obtained through K-means clustering on the nuPlan dataset~\citep{nuplan}. Following~\citep{li2024hydra, li2025hydra, yao2025drivesuprim}, we default to use the DD3D-pretrained~\citep{park2021pseudo} V2-99~\citep{lee2019energy} as the image backbone in our experiments. The ViT-L~\cite{dosovitskiy2020image} backbone is pretrained from Depth-Anything~\citep{yang2024depth}.

\input{tables/table1_navhard}

\subsection{Main Results}

\textbf{Navhard} is a difficult long-tail benchmark as its distribution widely differs from training split distribution. Thus, overcoming this distribution mismatch with strong dataset performance highlights how a method can generalize from one distribution to another. In ~\cref{table:navhard}, we quantitatively that ZTRS' dense supervision with EPO can generalize to even Navhard's distribution containing safety-critical scenarios. \hydra achieves SOTA performance across various settings. Equipped with V2-99 and EVA-ViT-L backbones, \hydra significantly outperforms the strong GTRS-Dense baseline by +3.8 EPDMS and +2.8 EPDMS, respectively. In particular, these gains are achieved despite the use of additional regularization and dynamic trajectory candidates in GTRS-Dense, which are not required by our framework.

\textbf{HUGSIM} is a safety-critical closed-loop dataset where human demonstrations are not available. In ~\cref{tab:hugsim}, we show how ZTRS, even trained with open-loop rewards, can transfer successfully to a closed-loop setting and overcome the training lack of human demonstrations. \hydra demonstrates superior robustness in closed-loop, achieving SOTA results on the HUGSIM benchmark with an overall 42.6 RC and 28.9 HD-Score. Notably, these results are achieved through zero-shot closed-loop transfer, as the model is trained purely on the open-loop Navtrain dataset without exposure to HUGSIM's synthetic interactive scenarios. 
Achieving better closed-loop performance than its IL baseline GTRS-Dense indicates that \hydra is not learning to ``hack'' open-loop rewards through EPO. Instead, EPO facilitates generalization to closed-loop settings.

\input{tables/table2_hugsim}
\input{tables/table3_navsim}

\textbf{Navtest.} Finally, we show in ~\cref{tab:navsim} that \hydra performs on par with SOTA IL planners in nominal driving scenarios found in Navtest. We note we are able to achieve SOTA despite the fact that reported metrics include ego progress (EP), which measures the L2 distance to the expert trajectories, none of which we ever trained with. Thus, we see a notable gap in our EP score compared to the IL methods. Even so, with a lightweight ResNet34 backbone, \hydra achieves an EPDMS of 83.2, outperforming models with advanced trajectory scoring techniques like DriveSuprim~\cite{yao2025drivesuprim}. With higher-capacity backbones like ViT-L and V2-99, \hydra maintains competitive results (86.2 EPDMS and 85.3 EPDMS) against DriveSuprim and surpasses its IL baseline GTRS-Dense~\cite{li2025generalized} by a clear margin (+1.5 EPDMS and +1.3 EPDMS).

\input{figs/curves}
\input{tables/abaltion1}

\subsection{Ablation Studies}

\textbf{RL Policy Gradient Comparisons.} We revisit the common cold-start RL problem, where policy gradient methods struggle to find high-rewards in an large continuous space without proper guidance. This is especially visible in early training stages. As such, we analyze our EPO against other policy gradient methods in the beginning epochs, as seen in ~\cref{fig:curves}. We evaluate against three policy methods within a discrete trajectory scoring framework: REINFORCE~\cite{sutton1999policy} and two common settings of known high-performer GRPO~\cite{shao2024deepseekmath}. To ensure fair comparison, all baselines sample from the same trajectory vocabulary. In ~\cref{fig:curves}, we observe that EPO demonstrates stable, non-volatile performance on both Navhard and Navtest. Its strong initial performance illustrates how EPO's dense supervision enables us to successfully find high-reward trajectories. Conversely, REINFORCE shows noticeably weaker performance, especially on Navtest. Importantly, although GRPO is widely acknowledged for its performance delivery, we not only see both GRPO settings' poor performance but also their instability through their performance fluctuations, which is undesirable due to potential model collapse. This is likely due to GRPO's unrepresentative subset trajectory sampling and the prevalence of negative gradients in sampled trajectory groups.

\textbf{Learning Paradigms Comparison.} In Tab.~\ref{table:abl_learning}, we study the effects of different learning paradigms and targets.
When using the trajectory with the maximum EPDMS score (i.e., $\hat{\mathcal{E}}=\underset{a}{\text{argmax}}\space\mathcal{E}(s,a)$) as the imitation target, performance drops significantly compared to the IL baseline, as many trajectories in $\mathcal{A}$ can achieve high EPDMS and a single target fails to capture the underlying pattern.
Using likelihoods over the entire action space mitigates this issue, improving EPDMS by 7.5\%, but introduces serious oscillation, as indicated by the low EC metric.
This highlights the need for our correction term $b$ to enforce temporal consistency. Using $\mathcal{E}-b$ as the reward increases EC by 23.4\%.

\input{tables/abaltion2}
\input{tables/ablation3_seed}

\textbf{Action Space Size Ablation. }
Tab.~\ref{table:abl_vocab} shows the relationship between the size of the action space and evaluation data. Models using the full action space during inference achieve the best results on real-world data, as reflected by EPDMS on Navtest and Navhard, while shrinking the action space improves performance on the simulated portion. This finding is consistent with GTRS~\citep{li2025generalized}: regardless of whether the model is trained with human demonstrations, reducing model complexity enhances generalization on unseen simulated data.

\textbf{Sensitivity to Training Initialization.} Here, we evaluate the training stability and variance of our proposed Exhaustive Policy Optimization (EPO) against the classical policy gradient method, REINFORCE. As shown in Tab.~\ref{tab:seeds}, the REINFORCE baseline exhibits significant performance fluctuations across different initialization seeds (75.0 vs. 72.6 EPDMS), highlighting the randomness of stochastic sampling in a large trajectory space. In contrast, EPO achieves substantially higher performance and stability, yielding a marginal variance of 0.2 EPDMS. Since EPO performs policy updates by evaluating a comprehensive and fixed trajectory vocabulary exhaustively at each training step, it provides a more deterministic optimization landscape and reduces such fluctuations.

\input{figs/viz_navsim}

\vspace{-2mm}
\subsection{Qualitative Results}

\input{figs/viz_hugsim}

Fig. \ref{fig:viz_navsim} and Fig. \ref{fig:viz_hugsim} show visualization results on the open-loop Navtest and the closed-loop HUGSIM, respectively. Interestingly, although \hydra is trained without human demonstrations, \hydra learns driving patterns that resemble human trajectories from rule-based rewards in nominal driving scenarios, such as long-term path following. 
On the other hand, when facing complex urban interactions, as shown in Fig. \ref{fig:viz_navsim} (d), \hydra can crucially make different but safer trajectory plans than the human demonstration. In the red box's leftmost scenario, while both the human and the agent navigate around a cyclist, \hydra exhibits a more pronounced nudging behavior, providing even more lateral clearance than the human driver to ensure safety. Further, the second scenario shows that \hydra decelerates after detecting the cut-in behavior of the leading vehicle. In the third scenario, \hydra shows it can anticipate the luggage handlers' movements, opting for a more conservative path than that of the human driver. 

Moreover, \hydra manages to navigate safely in safety-critical driving scenarios without closed-loop training or adaptation, as shown in Fig. \ref{fig:viz_hugsim}. Even under the extreme conditions depicted in Fig. \ref{fig:viz_hugsim} (c), where the ego agent must overtake a parked car while facing an oncoming vehicle, \hydra can safely complete the route. This demonstrates the strong closed-loop driving ability and the robust safety margin maintained by \hydra in highly complex long-tail situations.

%% file: tables/table1_navhard.tex
\begin{table*}[!t]
\centering
\scriptsize
\caption{\textbf{Performance on the Navhard Benchmark.} ZTRS achieves SOTA results, demonstrating our ability to generalize to different distributions, including safety-critcal cases in Navhard. We note that PDM-Closed uses ground-truth symbolic inputs for planning. \textit{H-Demo: human demonstration}.}

\scriptsize
\resizebox{\textwidth}{!}{
\begin{tabular}{c |c|c |c c c c c c c c c }

    \toprule

    Backbone & Method & H-Demo & \multicolumn{2}{c|}{Inputs} & \multicolumn{4}{c|}{Planner Type} & \multicolumn{3}{c}{EPDMS $\uparrow$} \\
    \midrule
    
    - & PDM-Closed~\citep{dauner2023parting} & - & \multicolumn{2}{c|}{Privileged Info.} & \multicolumn{4}{c|}{Rule-based Planner} & \multicolumn{3}{c}{51.3} \\

    \midrule
    \multirow{3}{*}{ResNet34} & LTF~\citep{chitta2022transfuser} & \usym{2713} & \multicolumn{2}{c|}{Image} & \multicolumn{4}{c|}{Regression-based Planner} & \multicolumn{3}{c}{23.1} \\
    
    & DiffusionDrive~\citep{liao2025diffusiondrive} & \usym{2713}& \multicolumn{2}{c|}{LiDAR+Image} & \multicolumn{4}{c|}{Diffusion-based Planner} & \multicolumn{3}{c}{27.5} \\
    & \gc \hydra (Ours) & \gc  \usym{2717}& \multicolumn{2}{c|}{\gc Image} & \multicolumn{4}{c|}{\gc Trajectory Scorer} & \multicolumn{3}{c}{\gc \textbf{39.9}} \\
    
    \midrule
    \multirow{3}{*}{V2-99} & DriveSuprim~\citep{yao2025drivesuprim} & \usym{2713}& \multicolumn{2}{c|}{Image}& \multicolumn{4}{c|}{Trajectory Scorer} & \multicolumn{3}{c}{42.1} \\
    
     & GTRS-Dense~\citep{li2025generalized} & \usym{2713}& \multicolumn{2}{c|}{Image}& \multicolumn{4}{c|}{Trajectory Scorer} & \multicolumn{3}{c}{41.7} \\
    
     & \gc \hydra (Ours) & \gc \usym{2717}& \multicolumn{2}{c|}{\gc Image}& \multicolumn{4}{c|}{\gc Trajectory Scorer} & \multicolumn{3}{c}{\gc \textbf{45.5}}    \\

    \midrule
    \multirow{3}{*}{EVA-ViT-L} & DriveSuprim~\citep{yao2025drivesuprim} & \usym{2713}& \multicolumn{2}{c|}{Image}& \multicolumn{4}{c|}{Trajectory Scorer} & \multicolumn{3}{c}{44.7} \\
    
     & GTRS-Dense~\citep{li2025generalized} & \usym{2713}& \multicolumn{2}{c|}{Image}& \multicolumn{4}{c|}{Trajectory Scorer} & \multicolumn{3}{c}{43.4} \\
    
     & \gc \hydra (Ours) & \gc \usym{2717}& \multicolumn{2}{c|}{\gc Image}& \multicolumn{4}{c|}{\gc Trajectory Scorer} & \multicolumn{3}{c}{\gc \textbf{46.2}}    \\
     \midrule
     \addlinespace[4pt]
     \midrule
    
    Backbone 
    & Method
    & H-Demo
    & $\text{NC}$ $\uparrow$
    & $\text{DAC}$ $\uparrow$
    & $\text{DDC}$ $\uparrow$
    & $\text{TLC}$ $\uparrow$
    & $\text{EP}$ $\uparrow$
    & $\text{TTC}$ $\uparrow$
    & $\text{LK}$ $\uparrow$
    & $\text{HC}$ $\uparrow$
    & $\text{EC}$ $\uparrow$ \\
    \midrule
    \multicolumn{12}{c}{\textit{Stage 1: Real-world Scenarios}} \\
    \midrule
    - & PDM-Closed~\citep{dauner2023parting} & - & 94.4 & 98.8 & 100 & 99.5 & 100 & 93.5 & 99.3 & 87.7 & 36.0  \\
    \midrule
    \multirow{2}{*}{ResNet34} & LTF~\citep{chitta2022transfuser} & \usym{2713} & 97.3 & 80.2 & 97.8 & 99.3 & 83.4 & 96.2 & 92.9 & \textbf{97.8} & 71.1  \\
    & DiffusionDrive~\citep{liao2025diffusiondrive} & \usym{2713} & 96.8 & 86.0 & 98.8 & 99.3 & \textbf{84.0} & 95.8 & \textbf{96.7} & 97.6 & \textbf{79.6}  \\
    & \gc \hydra (Ours) & \gc \usym{2717} & \gc \textbf{98.4} & \gc \textbf{92.9} & \gc \textbf{99.3} & \gc \textbf{100.0} & \gc 62.8 & \gc \textbf{98.7} & \gc 93.3 & \gc 97.3 & \gc 43.1  \\
    
    \midrule
    
    \multirow{3}{*}{V2-99} & DriveSuprim~\citep{yao2025drivesuprim} & \usym{2713}  &
    98.9 & 95.1 & 99.2 & 99.6 & \textbf{76.1} & \textbf{99.1} & 94.7 & 97.6 & \textbf{54.2}  \\
    
     & GTRS-Dense~\citep{li2025generalized} & \usym{2713} & 98.7 & 95.8 & 99.4 & 99.3 & 72.8 & 98.7 & 95.1 & 96.9 & 40.4  \\
    
     & \gc \hydra (Ours) & \gc \usym{2717} & \gc  \textbf{98.9}  & \gc \textbf{97.6}  & \gc \textbf{100.0} & \gc \textbf{100.0} & \gc 66.7 & \gc 98.9 & \gc \textbf{96.2} & \gc 96.7 & \gc 44.0  \\
    \midrule

    \multirow{3}{*}{EVA-ViT-L} & DriveSuprim~\citep{yao2025drivesuprim} & \usym{2713} & 98.7 & 98.0 & 99.1 & 99.8 & 75.9 & 98.7 & 94.7 & \textbf{97.6} & \textbf{49.8}  \\
    
     & GTRS-Dense~\citep{li2025generalized} & \usym{2713}  & 97.6 & 95.8 & 99.7 & 99.8  & \textbf{77.2}  & 97.8  & \textbf{95.3}  & 97.3  & 46.7   \\
    
     & \gc \hydra (Ours) & \gc \usym{2717} & \gc \textbf{99.1}  & \gc \textbf{98.9} & \gc \textbf{100.0} & \gc \textbf{100.0} & \gc 65.6  & \gc \textbf{98.9} & \gc 94.4 & \gc 95.8 & \gc 38.2 \\

    \midrule 
    \addlinespace[4pt]
    \midrule
    Backbone 
    & Method
    & H-Demo
    & $\text{NC}$ $\uparrow$
    & $\text{DAC}$ $\uparrow$
    & $\text{DDC}$ $\uparrow$
    & $\text{TLC}$ $\uparrow$
    & $\text{EP}$ $\uparrow$
    & $\text{TTC}$ $\uparrow$
    & $\text{LK}$ $\uparrow$
    & $\text{HC}$ $\uparrow$
    & $\text{EC}$ $\uparrow$ \\
    \midrule
    \multicolumn{12}{c}{\textit{Stage 2: Synthetic Scenarios}} \\
    \midrule
    - & PDM-Closed~\citep{dauner2023parting} & - & 88.1 & 90.6 & 96.3 & 98.5 & 100 &  83.1 & 73.7 & 91.5 & 25.4 \\
    \midrule
    
    \multirow{2}{*}{ResNet34} & LTF~\citep{chitta2022transfuser} & \usym{2713} &
    79.4 & 69.0 & 85.6 & 98.5 & 83.8 & 76.7 & 47.9 & 97.0 & 70.6  \\
    
    & DiffusionDrive~\citep{liao2025diffusiondrive} & \usym{2713} & 80.1 & 72.8 & 84.4 & 98.4 & \textbf{85.9} & 76.6 & 46.4 & 96.3 & \textbf{72.8} \\
    
    & \gc \hydra (Ours) & \gc \usym{2717} & \gc \textbf{91.8} & \gc \textbf{88.3} & \gc \textbf{96.2} & \gc \textbf{98.9} & \gc 54.6 & \gc \textbf{90.2} & \gc \textbf{56.4} & \gc \textbf{98.2} & \gc 63.0  \\
    
    \midrule

    \multirow{3}{*}{V2-99} & DriveSuprim~\citep{yao2025drivesuprim} & \usym{2713}  &
     87.9 & 88.8 & 89.6 & 98.8 & \textbf{80.3} & 86.0 & 53.5 & 97.1 & 56.1 \\
    
     & GTRS-Dense~\citep{li2025generalized} & \usym{2713}  & \textbf{91.4} & 89.2 & 94.4 &  98.8 & 69.5 & \textbf{90.1} & 54.6 & 94.1 & 49.7  \\
    
     & \gc \hydra (Ours) & \gc \usym{2717} & \gc 91.1 & \gc \textbf{90.4} & \gc\textbf{95.8} & \gc \textbf{99.0} & \gc 63.6 & \gc  89.8 & \gc  \textbf{60.4} &  \gc \textbf{97.6} & \gc \textbf{66.1}   \\
    \midrule

    \multirow{3}{*}{EVA-ViT-L} & DriveSuprim~\citep{yao2025drivesuprim} & \usym{2713}  &  89.5 &  89.6 &  92.9 &  98.5 &  \textbf{78.9} &  86.4 &  55.3 & 96.5 & 52.7  \\
    
     & GTRS-Dense~\citep{li2025generalized} & \usym{2713} &  \textbf{91.9} & 91.3 &  92.7 &  \textbf{99.0} &  72.7 &  90.4 &  53.8 & 94.1 &  41.6   \\
    
     &  \gc \hydra (Ours) & \gc \usym{2717} &  \gc 91.2 & \gc \textbf{94.4} & \gc \textbf{96.3} & \gc 98.7 & \gc  59.4 & \gc \textbf{90.5} & \gc \textbf{57.3} & \gc\textbf{97.5} & \gc \textbf{63.0}  \\

\bottomrule
\end{tabular}
}

\label{table:navhard}

\end{table*}

%% file: tables/table2_hugsim.tex
\begin{table*}[!t]
\centering

\caption{\textbf{Zero-shot Performance on the HUGSIM Benchmark.} UniAD and VAD are trained on nuScenes~\cite{caesar2020nuscenes}, which partially overlap with the evaluation scenarios. Other methods are evaluated with zero-shot transfer.
\textit{H-Demo: human demonstration}.}
\resizebox{\textwidth}{!}{
\begin{tabular}{c|c|c|cccccccc|cc}
\toprule
\multirow{2}{*}[-2pt]{Method} 
& \multirow{2}{*}[-2pt]{H-Demo} 
& \multirow{2}{*}[-2pt]{Zero-shot}
& \multicolumn{2}{c}{Easy} 
& \multicolumn{2}{c}{Medium} 
& \multicolumn{2}{c}{Hard} 
& \multicolumn{2}{c|}{Extreme} 
& \multicolumn{2}{c}{Overall} \\
\cmidrule{4-13}
 & & & RC $\uparrow$ & HD-Score $\uparrow$ & RC $\uparrow$ & HD-Score $\uparrow$ & RC $\uparrow$ & HD-Score $\uparrow$ & RC $\uparrow$ & HD-Score $\uparrow$ & RC $\uparrow$ & HD-Score $\uparrow$ \\
\midrule

VAD~\citep{jiang2023vad} & \usym{2713} & \usym{2717} 
& 38.7 & 24.3 & 27.0  & 9.9  & 25.5  & 10.4  & 23.0  & 8.2  &  27.9 & 12.3  \\

UniAD~\citep{hu2023planning}& \usym{2713} & \usym{2717} 
& 58.6 & 48.7 &  41.2 &  29.5 & \textbf{40.4}  & \textbf{27.3}  & \textbf{26.0}  &  \textbf{14.3} & 40.6  &  28.9 \\

LTF~\citep{chitta2022transfuser} & \usym{2713} & \usym{2713} 
& 68.4 & 52.8 & 40.7  & 24.6  & 36.9  &  19.8 & 25.5  &  8.1 &  41.4 &  24.8 \\

GTRS-Dense~\citep{li2025generalized} & \usym{2713} & \usym{2713} 
& 62.0 & 57.8  & 40.5 & 35.2 & 23.9  &  16.2 & 13.7  & 5.8  & 34.5  & 28.2 \\

\rowcolor{gray!15}
\hydra (Ours) & \usym{2717} & \usym{2713} 
& \textbf{74.8}  &  \textbf{66.9} &  \textbf{46.1} &  \textbf{37.9} & 31.6 & 21.1  &  20.0 & 9.8  &  \textbf{42.0} &  \textbf{32.9} \\
\bottomrule
\end{tabular}
}

\label{tab:hugsim}
\end{table*}

%% file: tables/table3_navsim.tex
\begin{table*}[!t]
    \centering
\caption{\textbf{Performance on the Navtest Benchmark.} \hydra performs on par with SOTA IL planners in nominal scenarios. Notably, while Imitation Learning methods relying on huamn demonstration (H-Demo) achieve higher Ego Progress (EP) by mimicking human trajectories, \hydra remains highly competitive across most metrics despite never being trained on such trajectories.}
    \small
    \resizebox{\textwidth}{!}{
    \begin{tabular}{c|c|c|ccccccccc|c}
        \toprule
        Backbone & Method & H-Demo & NC $\uparrow$ & DAC $\uparrow$ & DDC $\uparrow$ & TL $\uparrow$ & EP $\uparrow$ & TTC $\uparrow$ & LK $\uparrow$ & HC $\uparrow$ & EC $\uparrow$ & EPDMS $\uparrow$ \\
        \midrule
        - & Human Agent & - & 100 & 100 & 99.8 & 100 & 87.4 & 100 & 100 & 98.1 & 90.1 & 90.3 \\
        \midrule
        \midrule
        - & Ego Status MLP & \usym{2713} & 93.1 & 77.9 & 92.7 & 99.6 & 86.0 & 91.5 & 89.4 & 98.3 & 85.4 & 64.0 \\
        \midrule

        \multirow{4}{*}{ResNet34} & Transfuser~\citep{chitta2022transfuser} & \usym{2713} & 96.9 & 89.9 & 97.8 & 99.7 & 87.1 & 95.4 & 92.7 & \textbf{98.3} & \textbf{87.2} & 76.7 \\
        & HydraMDP++~\citep{li2025hydra} & \usym{2713} & 97.2 & 97.5 & 99.4 & 99.6 & 83.1 & 96.5 & 94.4 & 98.2 & 70.9 & 81.4 \\
        & DriveSuprim~\citep{yao2025drivesuprim} & \usym{2713} & 97.5 & 96.5 & 99.4 & 99.6 & \textbf{88.4} & 96.6 & \textbf{95.5} & \textbf{98.3} & 77.0 & 83.1 \\
        & \gc \hydra (Ours) & \gc \usym{2717} & \gc \textbf{97.9} & \gc \textbf{98.0} & \gc \textbf{99.7} & \gc \textbf{99.8} & \gc 80.1 & \gc \textbf{96.9} & \gc 94.9 & \gc 98.1 & \gc 79.0 & \gc \textbf{83.2} \\
        \midrule

        \multirow{4}{*}{ViT-L} & HydraMDP++~\citep{li2025hydra} & \usym{2713} & 98.5 & 98.5 & 99.5 & 99.7 & 87.4 & 97.9 & 95.8 & 98.2 & 75.7 & 85.6 \\
        & DriveSuprim~\citep{yao2025drivesuprim} & \usym{2713} & 98.4 & 98.6 & 99.6 & 99.8 & \textbf{90.5} & 97.8 & \textbf{97.0} & \textbf{98.3} & \textbf{78.6} & \textbf{87.1} \\
        & GTRS-Dense~\citep{li2025generalized} & \usym{2713} & \textbf{99.0} & 97.8 & 99.3 & \textbf{99.9} & 86.3 & \textbf{98.3} & 95.1 & \textbf{98.3} & 71.9 & 84.7 \\
        & \gc \hydra (Ours) & \gc \usym{2717} & \gc 98.2 & \gc \textbf{99.1} & \gc \textbf{99.7} & \gc 99.8 & \gc 86.9 & \gc 97.5 & \gc 96.6 & \gc 98.2 & \gc 78.2 & \gc 86.2 \\
        \midrule
        
        \multirow{4}{*}{V2-99} & HydraMDP++~\citep{li2025hydra} & \usym{2713} & \textbf{98.4} & 98.0 & 99.4 & 99.8 & 87.5 & \textbf{97.7} & 95.3 & \textbf{98.3} & 77.4 & 85.1 \\
        & DriveSuprim~\citep{yao2025drivesuprim} & \usym{2713} & 97.8 & 97.9 & 99.5 & \textbf{99.9} & \textbf{90.6} & 97.1 & 96.6 & \textbf{98.3} & \textbf{77.9} & \textbf{86.0} \\
        & GTRS-Dense~\citep{li2025generalized} & \usym{2713} & 97.6 & 98.5 & 99.5 & \textbf{99.9} & 89.5 & 97.2 & \textbf{96.8} & 97.2 & 57.2 & 84.0 \\
        & \gc \hydra (Ours) & \gc \usym{2717} & \gc 97.8 & \gc \textbf{99.4} & \gc \textbf{99.8} & \gc 99.8 & \gc 84.1 & \gc 97.0 & \gc 96.2 & \gc 98.2 & \gc 77.2 & \gc 85.3 \\
        
        \bottomrule
    \end{tabular}
    }
    \label{tab:navsim}
\end{table*}

%% file: figs/curves.tex
\begin{figure*}[!t]
    \centering
    \includegraphics[width=0.9\linewidth]{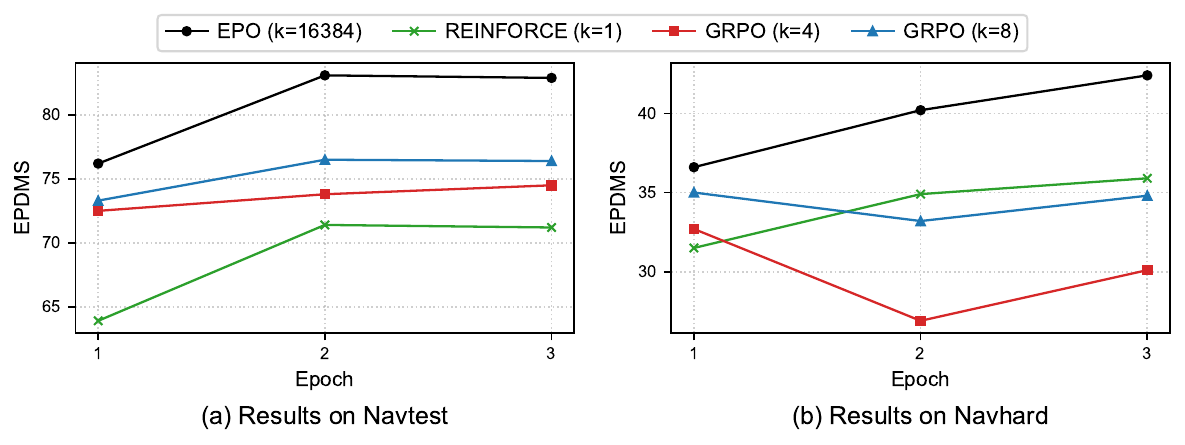}

    \caption{\textbf{Comparisons on Different RL Policy Gradient Methods.} k: Group size for sampled trajectories per training iteration. EPO enables stable optimization on both nominal driving (Navtest) and long-tail cases (Navhard). EPO's early out performance against baselines indicate that we are able to find high-reward trajectories early on, illustrating our ability to mitigate the cold-start problem. }

    \label{fig:curves}
\end{figure*}

%% file: tables/abaltion1.tex
\begin{table*}[t]
\scriptsize
\centering

\caption{\textbf{Ablation study on different learning methods and targets. } $\hat{\mathcal{E}}$ represents using the trajectory with the maximum ground-truth EPDMS as the imitation target. 
The Extended Comfort (EC) metric highlights that incorporating the correction term $b$ into the reward function is essential for achieving temporal consistency.
}
\resizebox{\textwidth}{!}{
\begin{tabular}{c c c| c c c c c c c c c | c}

    \toprule
    IL & RL & Target 

    & {NC} $\uparrow$
    & {DAC} $\uparrow$
    & {DDC} $\uparrow$
    & {TL} $\uparrow$
    & {EP} $\uparrow$
    & {TTC} $\uparrow$
    & {LK} $\uparrow$
    & {HC} $\uparrow$
    & {EC} $\uparrow$
    & {EPDMS} $\uparrow$ \\
    \midrule
    
    \usym{2713} & \usym{2717} & Human & 98.5 & 98.7 & 98.9 & 99.9 & 88.5 & 98.2 & 97.0 & 98.3 & 80.5 & 86.2 \\
    \midrule
    
    \usym{2713} & \usym{2717} & $\hat{\mathcal{E}}$ & 96.6 & 96.8 & 99.5 & 99.6 & 88.3 & 96.0 & 96.7 & 92.2 & \gc 18.5 & 76.7 \\

    \usym{2717} & \usym{2713} & $\mathcal{E}$ & 97.5 & 99.2 & \textbf{99.8} & 99.7 & \textbf{89.3} & 96.9 & \textbf{96.8} & 98.0 & \gc 53.8 & 84.2 \\
    \usym{2717} & \usym{2713} & $\mathcal{E}-b$ & \textbf{97.8} & \textbf{99.4} & \textbf{99.8} & \textbf{99.8} & 84.1 & \textbf{97.0} & 96.2 & \textbf{98.2} & \gc \textbf{77.2} & \textbf{85.3} \\
    
\bottomrule
\end{tabular}
}

\label{table:abl_learning}
\end{table*}

%% file: tables/abaltion2.tex
\begin{table*}[!t]
\scriptsize
\centering
\caption{\textbf{The relationship between the size of the action space and evaluation data.} $\text{EPDMS}_{1}$ measures the Stage 1 real-world scenarios of Navhard, while $\text{EPDMS}_{2}$ measures the Stage 2 simulated scenarios.}
\resizebox{0.9\textwidth}{!}{
\begin{tabular}{c|cc|c|ccc}
    \toprule
    \multirow{2}{*}[-2pt]{Backbone}
    & \multirow{2}{*}[-2pt]{$|\mathcal{A}|$ for training}
    & \multirow{2}{*}[-2pt]{$|\mathcal{A}|$ for inference}
    & Navtest & \multicolumn{3}{c}{Navhard} \\
    \cmidrule{4-7}
    & & 
    & EPDMS $\uparrow$
    & $\text{EPDMS}_{1}$ $\uparrow$
    & $\text{EPDMS}_{2}$ $\uparrow$
    & $\text{EPDMS}$ $\uparrow$  \\
    \midrule
    \multirow{3}{*}{V2-99} & 8192 & 8192  & 84.6 & 73.3 & 57.4 & 43.0 \\
    & 16384 & 16384     & \textbf{85.3} & \textbf{74.9} & 57.1 & 43.4\\
    & 16384 & 8192     & 82.0 & 74.2 & \textbf{60.7} & \textbf{45.5} \\
    \midrule
    \multirow{3}{*}{ViT-L} & 8192 & 8192  & 84.6 & 73.7 & 55.9 & 41.9 \\
     
    & 16384 & 16384 & \textbf{86.2} & \textbf{76.1} & 50.5 & 38.8 \\
    
    & 16384 & 8192  & 84.3 & 73.4 & \textbf{59.9} & \textbf{45.0} \\
\bottomrule
\end{tabular}
}

\label{table:abl_vocab}
\end{table*}

%% file: tables/ablation3_seed.tex
\begin{table}[t]
\centering
\caption{\textbf{Planning Performance as a Function of Training Initialization on Navtest.} Results across different initialization seeds show that our Exhaustive Policy Optimization (EPO) exhibits significantly lower variance and higher overall performance, indicating that our offline RL approach is stable and robust to initialization.}
\resizebox{0.7\textwidth}{!}{
\begin{tabular}{c|c|c|c}
\toprule
\textbf{Policy Optimization Method} & \textbf{Backbone} & \textbf{Seed} & \textbf{EPDMS $\uparrow$} \\ \midrule
\multirow{2}{*}{REINFORCE} & \multirow{2}{*}{V2-99} & 0 &  \textbf{75.0} \\
 &   & 1 &  72.6 \\
 \midrule
\multirow{2}{*}{EPO (ours)} & \multirow{2}{*}{V2-99} & 0 & 85.3 \\
 &  & 1 & \textbf{85.5} \\

\bottomrule
\end{tabular}
}
\label{tab:seeds}
\end{table}

%% file: figs/viz_navsim.tex
\begin{figure*}[!t]
    \centering

    \includegraphics[width=\linewidth]{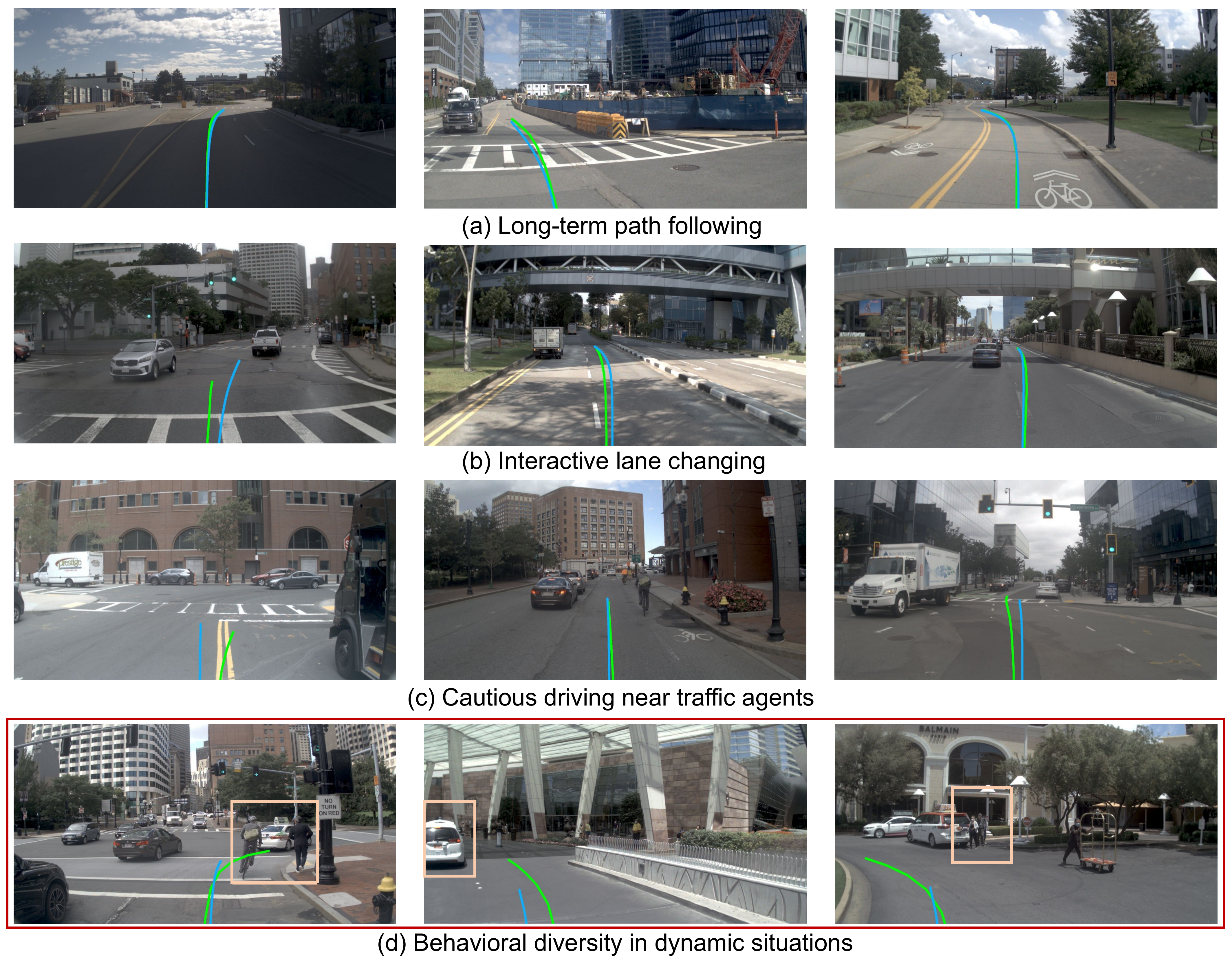}
    \caption{\textbf{Visualizations of \textcolor{cyan}{planned trajectories} and the \textcolor{green}{human trajectory} on Navtest.} In nominal scenarios, \hydra plans similarly to humans, exhibiting long-term path following. In complex cases, \hydra demonstrates sophisticated behaviors such as interactive lane changing, cautious navigation near traffic agents, and behavioral diversity in dynamic situations where it plans differently but more safely than humans.}
    
    \label{fig:viz_navsim}
\end{figure*}

%% file: figs/viz_hugsim.tex
\begin{figure*}[!t]
    \centering

    \includegraphics[width=\linewidth]{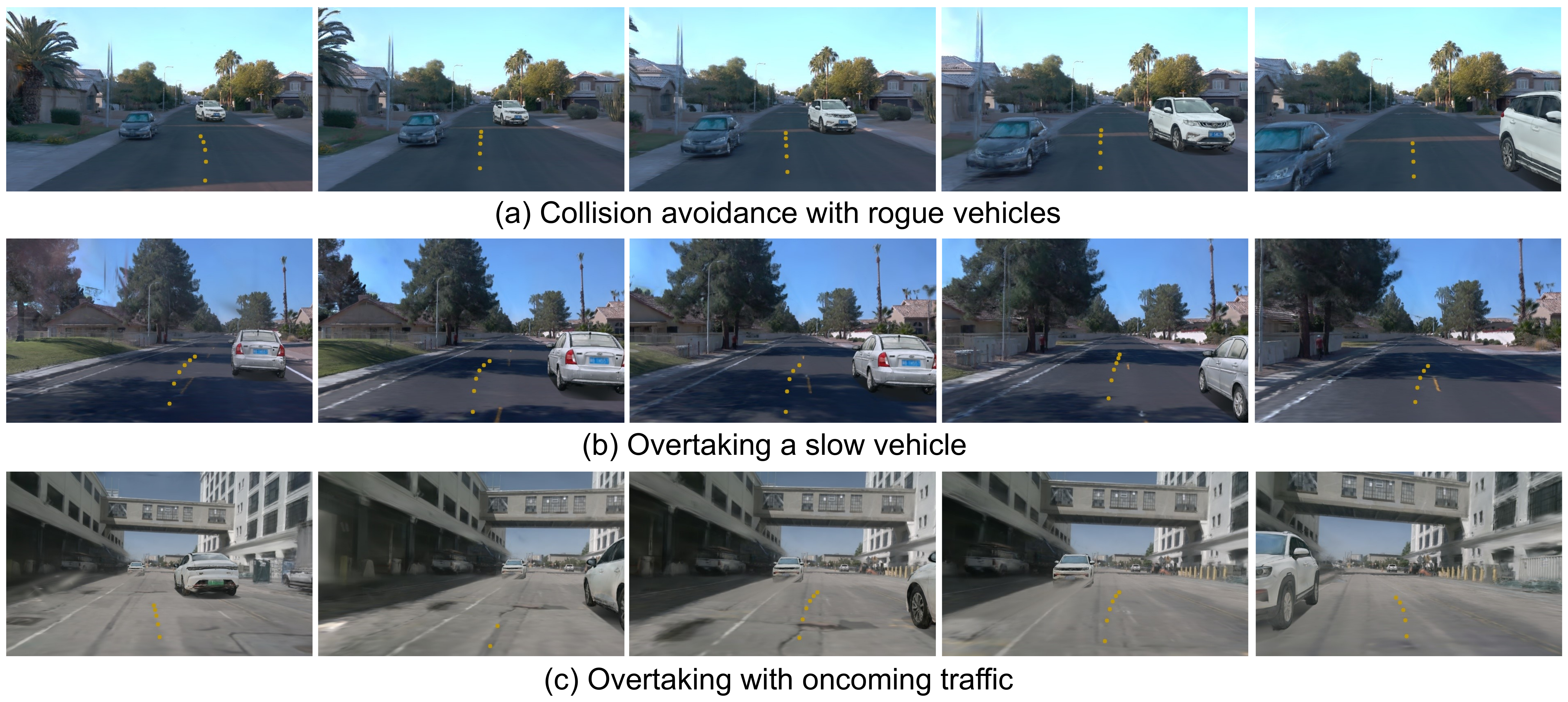}
    \caption{\textbf{Visualizations of \textcolor{orange}{planned trajectories} on the challenging closed-loop driving benchmark HUGSIM.} Here, we illustrate ZTRS can plan trajectories even in the shown long-tail scenarios, including safety critical, collision avoidance cases.}
    \label{fig:viz_hugsim}
\end{figure*}

%% file: sec/5_conclusion.tex
\section{Conclusion}
We proposed ZTRS, an effective end-to-end autonomous driving paradigm that eliminates the need for human demonstrations during training. By relying solely on offline data, reward-driven supervision, and a newly proposed exhaustive policy optimization (EPO), ZTRS successfully addresses the cold-start problem while enabling robust planning directly from high-dimensional sensor inputs, without human demonstrations. Our approach achieves strong SOTA performance on long-tail driving benchmarks such as NAVHARD, while maintaining top-level results on standard (nominal) driving benchmarks such as NAVSIM and HUGSIM. These findings demonstrate that well-designed reward mechanisms can potentially replace human demonstrations in training reliable end-to-end autonomous driving systems.

\textbf{Limitations and Discussions.} Despite \hydra achieving state-of-the-art performance on challenging planning and closed-loop driving benchmarks, it still lags behind imitation-based baselines in certain open-loop scenarios. EPO is also most direct when the planner acts over a fixed, enumerable trajectory vocabulary, since this allows rewards to be pre-computed and reused during training. Extending the same idea to continuous action spaces is therefore an important direction for future work. One practical route is to connect continuous trajectory generators with the discrete reward table through nearest-neighbor lookup, as suggested by recent continuous end-to-end planners~\citep{yao2026had}.

\textbf{Acknowledgements.} This work is supported by the National Natural Science Foundation of China (Grant No. 62427819) and the Science and Technology Commission of Shanghai Municipality (No. 24511103100).

\clearpage

%% file: supp_content.tex
\title{Zero-Human Demonstration End-to-end Autonomous Driving with Trajectory Scorer} 

\titlerunning{Zero-Human Demonstration End-to-end Autonomous Driving}

\authorrunning{Z. Li et al.}

\begin{center}
    {\Large\textbf{Zero-Human Demonstration End-to-end Autonomous Driving with Trajectory Scorer\\[2mm] Supplementary Material}}
\end{center}
\vspace{2mm}

\section{Architecture Details}

\begin{table}[h]
\centering
\caption{\textbf{Network Architecture Details.} TF $\times n$ denotes $n$ stacked Transformer layers. The Transformer encoder and decoder share the same hidden dimension, number of attention heads, and feed-forward network configuration.}
\resizebox{\textwidth}{!}{
\begin{tabular}{l|l}
\toprule
\textbf{Component} & \textbf{Implementation} \\ \midrule
Image backbone & ResNet34, V2-99, or EVA-ViT-L; Fully fine-tuned during RL training \\
Trajectory tokenizer & MLP (280$\rightarrow$1024$\rightarrow$256) + TF encoder $\times$1 \\
Trajectory decoder & TF decoder $\times$3 (8 heads, FFN 256$\rightarrow$1024$\rightarrow$256) \\
Scoring heads & MLP (256$\rightarrow$1024$\rightarrow$1) \\
Policy head & MLP (256$\rightarrow$1024$\rightarrow$1024$\rightarrow$1) \\
\bottomrule
\end{tabular}
}
\label{tab:network_arch}
\end{table}

\noindent We provide additional architectural details of ZTRS in Tab.~\ref{tab:network_arch}. The image backbone is fully fine-tuned during RL training rather than frozen. For the trajectory branch, each candidate trajectory is represented by 40 timesteps and 7 per-step features, including $(x,y,\text{heading})$ and motion-difference features, resulting in a 280-D input vector. This vector is encoded by an MLP and a single Transformer encoder layer to produce a 256-D trajectory token.

The trajectory tokens serve as queries in the Transformer decoder and attend to image tokens extracted by the backbone. The resulting 256-D attended trajectory features are passed to two types of prediction heads. The policy head outputs logits over the discrete trajectory vocabulary, while the scoring heads predict rule-based driving metric scores for each trajectory. Both heads operate independently on each trajectory token and produce one scalar per candidate, which enables ZTRS to score the entire vocabulary and apply Exhaustive Policy Optimization over all candidate trajectories.

\section{Ablations on Different Reward Components}
\begin{table*}[h]
\scriptsize
\centering
\caption{
\textbf{Planning Performance as a Function of Reward Components.} We group the EPDMS rewards into three categories — Safety (NC, DAC, DDC, TL, TTC, and LK), Progress (EP), and Comfort (HC and EC) — and analyze their impact on planning performance when included as training rewards. As shown, incorporating all reward categories enables our approach to prioritize safety while achieving smoother and more goal-directed behavior. }
\label{table:abl_rewards}
\resizebox{\textwidth}{!}{
\begin{tabular}{c c c| c c c c c c c c c | c}

    \toprule
    Safety & Progress & Comfort 

    & {NC} $\uparrow$
    & {DAC} $\uparrow$
    & {DDC} $\uparrow$
    & {TL} $\uparrow$
    & {EP} $\uparrow$
    & {TTC} $\uparrow$
    & {LK} $\uparrow$
    & {HC} $\uparrow$
    & {EC} $\uparrow$
    & {EPDMS} $\uparrow$ \\
    \midrule
    
    \usym{2713} & \usym{2717} & \usym{2717} & \textbf{98.6} & 99.3 & \textbf{99.9} & 99.7 & 63.3 & 96.3 & \textbf{96.2} & 79.8 & 24.5 & 72.0 \\
    
    \usym{2713} & \usym{2713} & \usym{2717} & 98.4 & 99.2 & 99.8 & 99.7 & 83.8 & \textbf{97.8} & 95.9 & 96.0 & 49.8 & 82.4 \\
    
    \usym{2713} & \usym{2713} & \usym{2713} & 97.8 & \textbf{99.4} & 99.8 & \textbf{99.8} & \textbf{84.1} & 97.0 & \textbf{96.2} & \textbf{98.2} & \textbf{77.2} & \textbf{85.3} \\
    
\bottomrule
\end{tabular}
}
\end{table*}

\noindent EPDMS is a rule-based metric that aggregates several driving-related dimensions, including time-to-collision, comfort, and lane keeping. These metrics can be grouped into three main categories: safety (NC, DAC, DDC, TL, TTC, and LK), progress (EP), and comfort (HC and EC).

In the first ablation study, we analyze the contribution of each category when used as a reward in our framework. To this end, the metrics are grouped into the three categories above, and we evaluate the final performance depending on whether each category is included or excluded from the reward. Tab.~\ref{table:abl_rewards} summarizes the results for this experiment.

A policy trained using only safety rewards achieves a baseline EPDMS score of 72.0. Although this configuration yields strong collision-avoidance performance (NC: 98.6), the low EP score reflects inefficient planning behavior. Adding the progress reward (EP), which is calculated by measuring each trajectory's progress along the rule-based PDM-Planner~\cite{dauner2023parting}'s route, substantially improves the overall score to 82.4. The progress reward encourages the model to explore trajectories that maintain ego velocity while respecting safety constraints. Finally, incorporating comfort rewards (HC and EC) further increases the performance to 85.3 EPDMS. This complete reward structure enables ZTRS to prioritize safety while also achieving smoother and more goal-directed behavior.

\section{Efficiency}

\begin{table}[h]
\centering

\caption{\textbf{Efficiency Comparison.} We compare the model complexity and inference speed of ZTRS against the GTRS-Dense~\cite{li2025generalized} baseline. FPS is measured on a single NVIDIA A100 GPU.}
\begin{tabular}{l|c|c|c}
\toprule
\textbf{Method} & \textbf{Backbone} & \textbf{Params.} & \textbf{FPS} \\ \midrule
GTRS-Dense~\cite{li2025generalized} & V2-99 & 83.0M & 18.5 \\
ZTRS & V2-99 & 83.4M & 18.4 \\
\bottomrule
\end{tabular}
\label{tab:efficiency}
\end{table}

\noindent We provide an efficiency comparison between ZTRS and GTRS-Dense~\cite{li2025generalized} in Tab.~\ref{tab:efficiency}. Since ZTRS is built upon a similar architecture as GTRS-Dense, the parameter count remains nearly identical (83.4M vs. 83.0M). Crucially, during the inference phase, when both methods utilize a standardized set of 8192 candidate trajectories, ZTRS achieves an inference speed of 18.4 FPS, which is indistinguishable from the baseline and well-suited for real-time driving. Regarding the training process, since the trajectory vocabulary is fixed, we pre-compute the rewards for each trajectory across the dataset in an offline manner. This strategy facilitates the efficiency of EPO without the need for redundant reward calculation.

\begin{figure*}[!t]
    \centering

    \includegraphics[width=\linewidth]{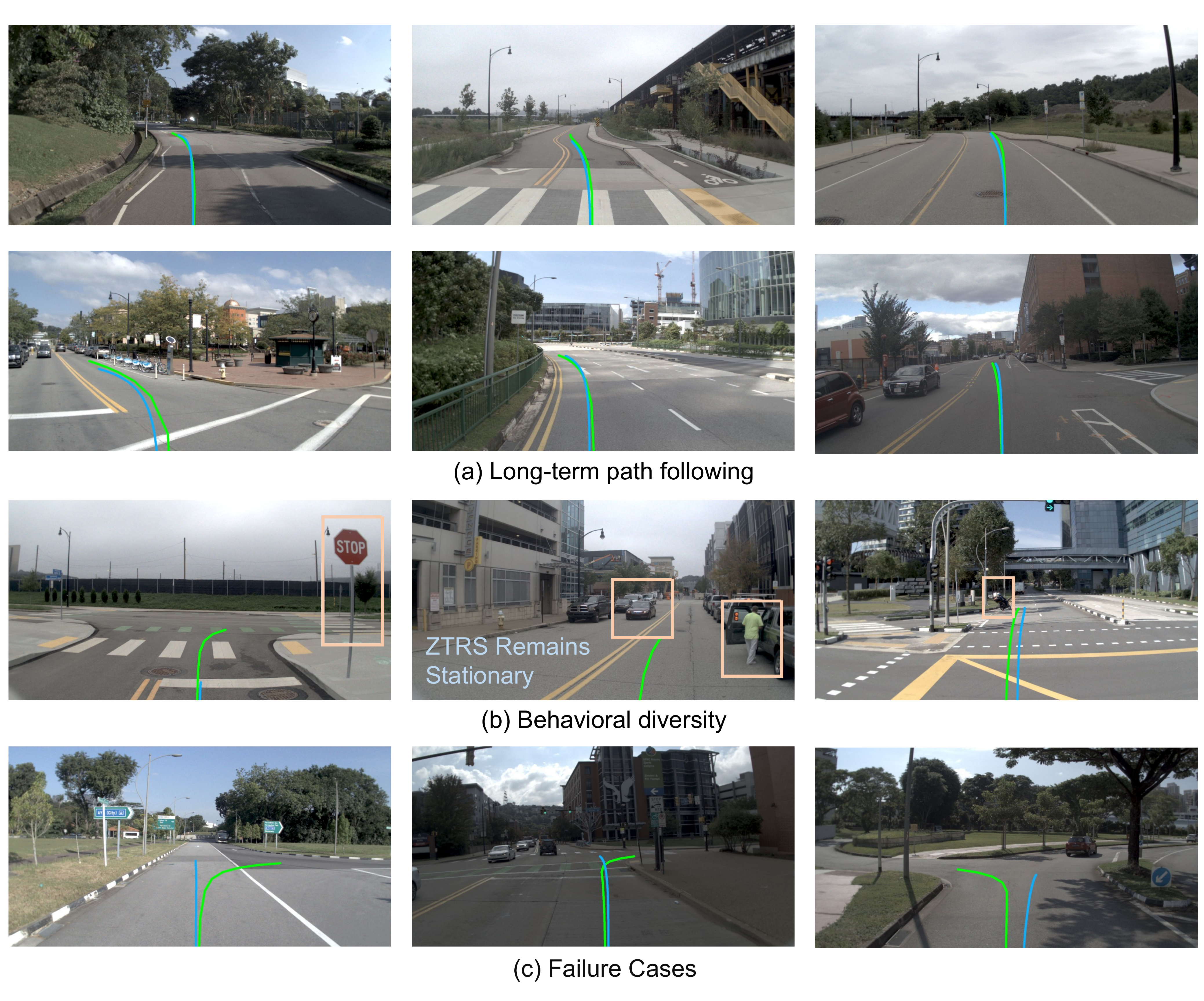}

    \caption{\textbf{Qualitative Results on the Navhard Benchmark.} We visualize the \textcolor{cyan}{planned trajectories} generated by ZTRS and the \textcolor{green}{human trajectories}. (a) demonstrates stable long-term path following across diverse urban environments. (b) showcases behavioral diversity, where ZTRS maintains large safety margins, such as remaining stationary at stop signs or in presence of pedestrians. (c) illustrates representative failure cases, primarily involving inconsistencies in command following .}
    
    \label{fig:viz_navsim_supp}
\end{figure*}

\section{Additional Qualitative Results}
Finally, in Fig.~\ref{fig:viz_navsim_supp}, we provide qualitative visualizations of our approach's driving behavior across different scenarios, including representative failure cases.
As shown in Fig.~\ref{fig:viz_navsim_supp}(a), our model demonstrates robust long-term path following, maintaining precise lane centering and smooth curvature even in complex road geometries. Furthermore, Fig.~\ref{fig:viz_navsim_supp}(b) highlights the behavioral diversity of our approach. For example, the model correctly identifies stop-controlled intersections and dynamic obstacles (e.g., pedestrians and motorcycles), choosing to remain stationary or maintain a larger safety margin than the human trajectory. As a representative failure example, in Fig.~\ref{fig:viz_navsim_supp} (c), ZTRS can occasionally exhibit inconsistent command following such as executing a path that deviates from the high-level navigational goal. This is, however, a problem common to other trajectory scorers~\cite{li2024hydra, li2025generalized, yao2025drivesuprim, wu2025navihydra}, as discussed in NaviHydra~\cite{wu2025navihydra}. Addressing this gap by aligning the zero-human reward structures with driving commands could be a promising direction for future work.

%% file: main.bib
@String(PAMI = {IEEE Trans. Pattern Anal. Mach. Intell.})

@String(CVPR= {IEEE Conf. Comput. Vis. Pattern Recog.})

@String(ICCV= {Int. Conf. Comput. Vis.})

@String(ECCV= {Eur. Conf. Comput. Vis.})

@String(NIPS= {Adv. Neural Inform. Process. Syst.})

@String(PAMI  = {IEEE TPAMI})

@String(CVPR  = {CVPR})

@String(ICCV  = {ICCV})

@String(ECCV  = {ECCV})

@String(NIPS  = {NeurIPS})

@article{chitta2022transfuser,
  title={Transfuser: Imitation with transformer-based sensor fusion for autonomous driving},
  author={Chitta, Kashyap and Prakash, Aditya and Jaeger, Bernhard and Yu, Zehao and Renz, Katrin and Geiger, Andreas},
  journal={IEEE Transactions on Pattern Analysis and Machine Intelligence},
  year={2022},
  publisher={IEEE}
}

@article{chen2024vadv2,
  title={VADv2: End-to-End Vectorized Autonomous Driving via Probabilistic Planning},
  author={Chen, Shaoyu and Jiang, Bo and Gao, Hao and Liao, Bencheng and Xu, Qing and Zhang, Qian and Huang, Chang and Liu, Wenyu and Wang, Xinggang},
  journal={arXiv preprint arXiv:2402.13243},
  year={2024}
}

@inproceedings{hu2023planning,
  title={Planning-oriented autonomous driving},
  author={Hu, Yihan and Yang, Jiazhi and Chen, Li and Li, Keyu and Sima, Chonghao and Zhu, Xizhou and Chai, Siqi and Du, Senyao and Lin, Tianwei and Wang, Wenhai and others},
  booktitle=CVPR,
  pages={17853--17862},
  year={2023}
}

@article{yao2026had,
  title={HAD: Combining Hierarchical Diffusion with Metric-Decoupled RL for End-to-End Driving},
  author={Yao, Wenhao and Sun, Xinglong and Li, Zhenxin and Lan, Shiyi and Wang, Zi and Alvarez, Jose M and Wu, Zuxuan},
  journal={arXiv preprint arXiv:2604.03581},
  year={2026}
}

@book{sutton1998reinforcement,
  title={Reinforcement learning: An introduction},
  author={Sutton, Richard S and Barto, Andrew G and others},
  volume={1},
  number={1},
  year={1998},
  publisher={MIT press Cambridge}
}

@article{li2025hydra,
  title={Hydra-mdp++: Advancing end-to-end driving via expert-guided hydra-distillation},
  author={Li, Kailin and Li, Zhenxin and Lan, Shiyi and Xie, Yuan and Zhang, Zhizhong and Liu, Jiayi and Wu, Zuxuan and Yu, Zhiding and Alvarez, Jose M},
  journal={arXiv preprint arXiv:2503.12820},
  year={2025}
}

@article{wu2025navihydra,
  title={NaviHydra: Controllable Navigation-guided End-to-end Autonomous Driving with Hydra-distillation},
  author={Wu, Hanfeng and Steiner, Marlon and Schmidt, Michael and Marcos-Ramiro, Alvaro and Stiller, Christoph},
  journal={arXiv preprint arXiv:2512.10660},
  year={2025}
}

@article{li2025finetuning,
  title={Finetuning generative trajectory model with reinforcement learning from human feedback},
  author={Li, Derun and Ren, Jianwei and Wang, Yue and Wen, Xin and Li, Pengxiang and Xu, Leimeng and Zhan, Kun and Xia, Zhongpu and Jia, Peng and Lang, Xianpeng and others},
  journal={arXiv preprint arXiv:2503.10434},
  year={2025}
}

@article{chang2021mitigating,
  title={Mitigating covariate shift in imitation learning via offline data with partial coverage},
  author={Chang, Jonathan and Uehara, Masatoshi and Sreenivas, Dhruv and Kidambi, Rahul and Sun, Wen},
  journal=NIPS,
  volume={34},
  pages={965--979},
  year={2021}
}

@article{li2025mtgs,
  title={MTGS: Multi-Traversal Gaussian Splatting},
  author={Li, Tianyu and Qiu, Yihang and Wu, Zhenhua and Lindstr{\"o}m, Carl and Su, Peng and Nie{\ss}ner, Matthias and Li, Hongyang},
  journal={arXiv preprint arXiv:2503.12552},
  year={2025}
}

@inproceedings{li2024think2drive,
  title={Think2drive: Efficient reinforcement learning by thinking with latent world model for autonomous driving (in carla-v2)},
  author={Li, Qifeng and Jia, Xiaosong and Wang, Shaobo and Yan, Junchi},
  booktitle=ECCV,
  pages={142--158},
  year={2024},
  organization={Springer}
}

@article{cusumano2025robust,
  title={Robust autonomy emerges from self-play},
  author={Cusumano-Towner, Marco and Hafner, David and Hertzberg, Alex and Huval, Brody and Petrenko, Aleksei and Vinitsky, Eugene and Wijmans, Erik and Killian, Taylor and Bowers, Stuart and Sener, Ozan and others},
  journal={arXiv preprint arXiv:2502.03349},
  year={2025}
}

@article{levine2020offline,
  title={Offline reinforcement learning: Tutorial, review, and perspectives on open problems},
  author={Levine, Sergey and Kumar, Aviral and Tucker, George and Fu, Justin},
  journal={arXiv preprint arXiv:2005.01643},
  year={2020}
}

@inproceedings{zhang2021end,
  title={End-to-end urban driving by imitating a reinforcement learning coach},
  author={Zhang, Zhejun and Liniger, Alexander and Dai, Dengxin and Yu, Fisher and Van Gool, Luc},
  booktitle=ICCV,
  pages={15222--15232},
  year={2021}
}

@article{jaeger2025carl,
  title={Carl: Learning scalable planning policies with simple rewards},
  author={Jaeger, Bernhard and Dauner, Daniel and Bei{\ss}wenger, Jens and Gerstenecker, Simon and Chitta, Kashyap and Geiger, Andreas},
  journal={arXiv preprint arXiv:2504.17838},
  year={2025}
}

@article{li2025hydranext,
  title={Hydra-next: Robust closed-loop driving with open-loop training},
  author={Li, Zhenxin and Wang, Shihao and Lan, Shiyi and Yu, Zhiding and Wu, Zuxuan and Alvarez, Jose M},
  journal={arXiv preprint arXiv:2503.12030},
  year={2025}
}

@article{yao2025drivesuprim,
  title={DriveSuprim: Towards Precise Trajectory Selection for End-to-End Planning},
  author={Yao, Wenhao and Li, Zhenxin and Lan, Shiyi and Wang, Zi and Sun, Xinglong and Alvarez, Jose M and Wu, Zuxuan},
  journal={arXiv preprint arXiv:2506.06659},
  year={2025}
}

@article{li2025generalized,
  title={Generalized Trajectory Scoring for End-to-end Multimodal Planning},
  author={Li, Zhenxin and Yao, Wenhao and Wang, Zi and Sun, Xinglong and Chen, Joshua and Chang, Nadine and Shen, Maying and Wu, Zuxuan and Lan, Shiyi and Alvarez, Jose M},
  journal={arXiv preprint arXiv:2506.06664},
  year={2025}
}

@article{wen2020fighting,
  title={Fighting copycat agents in behavioral cloning from observation histories},
  author={Wen, Chuan and Lin, Jierui and Darrell, Trevor and Jayaraman, Dinesh and Gao, Yang},
  journal=NIPS,
  volume={33},
  pages={2564--2575},
  year={2020}
}

@article{zhou2025autovla,
 title={AutoVLA: A Vision-Language-Action Model for End-to-End Autonomous Driving with Adaptive Reasoning and Reinforcement Fine-Tuning},
 author={Zhou, Zewei and Cai, Tianhui and Zhao, Seth Z.and Zhang, Yun and Huang, Zhiyu and Zhou, Bolei and Ma, Jiaqi},
 journal=NIPS,
 year={2025}
}

@article{noguchi2025offline,
  title={Offline Reinforcement Learning for End-to-End Autonomous Driving},
  author={Noguchi, Chihiro and Yamamoto, Takaki},
  journal={arXiv preprint arXiv:2512.18662},
  year={2025}
}

@inproceedings{nair2018overcoming,
  title={Overcoming exploration in reinforcement learning with demonstrations},
  author={Nair, Ashvin and McGrew, Bob and Andrychowicz, Marcin and Zaremba, Wojciech and Abbeel, Pieter},
  booktitle={ICRA},
  pages={6292--6299},
  year={2018},
  organization={IEEE}
}

@article{zou2025diffusiondrivev2,
  title={DiffusionDriveV2: Reinforcement learning-constrained truncated diffusion modeling in end-to-end autonomous driving},
  author={Zou, Jialv and Chen, Shaoyu and Liao, Bencheng and Zheng, Zhiyu and Song, Yuehao and Zhang, Lefei and Zhang, Qian and Liu, Wenyu and Wang, Xinggang},
  journal={arXiv preprint arXiv:2512.07745},
  year={2025}
}

@article{schulman2015high,
  title={High-dimensional continuous control using generalized advantage estimation},
  author={Schulman, John and Moritz, Philipp and Levine, Sergey and Jordan, Michael and Abbeel, Pieter},
  journal={arXiv preprint arXiv:1506.02438},
  year={2015}
}

@article{sutton1999policy,
  title={Policy gradient methods for reinforcement learning with function approximation},
  author={Sutton, Richard S and McAllester, David and Singh, Satinder and Mansour, Yishay},
  journal=NIPS,
  volume={12},
  year={1999}
}

@article{schulman2017proximal,
  title={Proximal policy optimization algorithms},
  author={Schulman, John and Wolski, Filip and Dhariwal, Prafulla and Radford, Alec and Klimov, Oleg},
  journal={arXiv preprint arXiv:1707.06347},
  year={2017}
}

@inproceedings{philion2020lift,
  title={Lift, splat, shoot: Encoding images from arbitrary camera rigs by implicitly unprojecting to 3d},
  author={Philion, Jonah and Fidler, Sanja},
  booktitle=ECCV,
  pages={194--210},
  year={2020},
  organization={Springer}
}

@inproceedings{phan2020covernet,
  title={Covernet: Multimodal behavior prediction using trajectory sets},
  author={Phan-Minh, Tung and Grigore, Elena Corina and Boulton, Freddy A and Beijbom, Oscar and Wolff, Eric M},
  booktitle=CVPR,
  pages={14074--14083},
  year={2020}
}

@inproceedings{dauner2023parting,
  title={Parting with misconceptions about learning-based vehicle motion planning},
  author={Dauner, Daniel and Hallgarten, Marcel and Geiger, Andreas and Chitta, Kashyap},
  booktitle={Conference on Robot Learning},
  pages={1268--1281},
  year={2023},
  organization={PMLR}
}

@INPROCEEDINGS{nuplan, 
  title={NuPlan: A closed-loop ML-based planning benchmark for autonomous vehicles},
  author={H. Caesar, J. Kabzan, K. Tan et al.},
  booktitle={CVPR ADP3 workshop},
  year=2021
}

@inproceedings{lee2019energy,
  title={An energy and GPU-computation efficient backbone network for real-time object detection},
  author={Lee, Youngwan and Hwang, Joong-won and Lee, Sangrok and Bae, Yuseok and Park, Jongyoul},
  booktitle={Proceedings of the IEEE/CVF conference on computer vision and pattern recognition workshops},
  pages={0--0},
  year={2019}
}

@inproceedings{Renz2022CORL,
    author       = {Katrin Renz and Kashyap Chitta and Otniel-Bogdan Mercea and A. Sophia Koepke and Zeynep Akata and Andreas Geiger},
    title        = {PlanT: Explainable Planning Transformers via Object-Level Representations},
    booktitle    = {CoRL},
    year         = {2022}
}

@article{yang2024depth,
  title={Depth anything: Unleashing the power of large-scale unlabeled data},
  author={Yang, Lihe and Kang, Bingyi and Huang, Zilong and Xu, Xiaogang and Feng, Jiashi and Zhao, Hengshuang},
  journal={arXiv preprint arXiv:2401.10891},
  year={2024}
}

@inproceedings{toromanoff2020end,
  title={End-to-end model-free reinforcement learning for urban driving using implicit affordances},
  author={Toromanoff, Marin and Wirbel, Emilie and Moutarde, Fabien},
  booktitle=CVPR,
  pages={7153--7162},
  year={2020}
}

@article{gulino2023waymax,
  title={Waymax: An accelerated, data-driven simulator for large-scale autonomous driving research},
  author={Gulino, Cole and Fu, Justin and Luo, Wenjie and Tucker, George and Bronstein, Eli and Lu, Yiren and Harb, Jean and Pan, Xinlei and Wang, Yan and Chen, Xiangyu and others},
  journal=NIPS,
  volume={36},
  pages={7730--7742},
  year={2023}
}

@article{vaswani2017attention,
  title={Attention is all you need},
  author={Vaswani, Ashish and Shazeer, Noam and Parmar, Niki and Uszkoreit, Jakob and Jones, Llion and Gomez, Aidan N and Kaiser, {\L}ukasz and Polosukhin, Illia},
  journal=NIPS,
  volume={30},
  year={2017}
}

@article{cao2025pseudo,
  title={Pseudo-simulation for autonomous driving},
  author={Cao, Wei and Hallgarten, Marcel and Li, Tianyu and Dauner, Daniel and Gu, Xunjiang and Wang, Caojun and Miron, Yakov and Aiello, Marco and Li, Hongyang and Gilitschenski, Igor and others},
  journal={CoRL},
  year={2025}
}

@inproceedings{shao2023safety,
  title={Safety-enhanced autonomous driving using interpretable sensor fusion transformer},
  author={Shao, Hao and Wang, Letian and Chen, Ruobing and Li, Hongsheng and Liu, Yu},
  booktitle={CoRL},
  pages={726--737},
  year={2023},
  organization={PMLR}
}

@article{hu2022model,
  title={Model-based imitation learning for urban driving},
  author={Hu, Anthony and Corrado, Gianluca and Griffiths, Nicolas and Murez, Zachary and Gurau, Corina and Yeo, Hudson and Kendall, Alex and Cipolla, Roberto and Shotton, Jamie},
  journal=NIPS,
  volume={35},
  pages={20703--20716},
  year={2022}
}

@inproceedings{jia2023think,
  title={Think twice before driving: Towards scalable decoders for end-to-end autonomous driving},
  author={Jia, Xiaosong and Wu, Penghao and Chen, Li and Xie, Jiangwei and He, Conghui and Yan, Junchi and Li, Hongyang},
  booktitle={CVPR},
  pages={21983--21994},
  year={2023}
}

@inproceedings{chen2020learning,
  title={Learning by cheating},
  author={Chen, Dian and Zhou, Brady and Koltun, Vladlen and Kr{\"a}henb{\"u}hl, Philipp},
  booktitle={CoRL},
  pages={66--75},
  year={2020},
  organization={PMLR}
}

@article{yang2025raw2drive,
  title={Raw2Drive: Reinforcement learning with aligned world models for end-to-end autonomous driving (in carla v2)},
  author={Yang, Zhenjie and Jia, Xiaosong and Li, Qifeng and Yang, Xue and Yao, Maoqing and Yan, Junchi},
  journal={arXiv preprint arXiv:2505.16394},
  year={2025}
}

@article{gao2025rad,
  title={Rad: Training an end-to-end driving policy via large-scale 3dgs-based reinforcement learning},
  author={Gao, Hao and Chen, Shaoyu and Jiang, Bo and Liao, Bencheng and Shi, Yiang and Guo, Xiaoyang and Pu, Yuechuan and Yin, Haoran and Li, Xiangyu and Zhang, Xinbang and others},
  journal={arXiv preprint arXiv:2502.13144},
  year={2025}
}

@inproceedings{jia2023driveadapter,
  title={Driveadapter: Breaking the coupling barrier of perception and planning in end-to-end autonomous driving},
  author={Jia, Xiaosong and Gao, Yulu and Chen, Li and Yan, Junchi and Liu, Patrick Langechuan and Li, Hongyang},
  booktitle=ICCV,
  pages={7953--7963},
  year={2023}
}

@inproceedings{park2021pseudo,
  title={Is pseudo-lidar needed for monocular 3d object detection?},
  author={Park, Dennis and Ambrus, Rares and Guizilini, Vitor and Li, Jie and Gaidon, Adrien},
  booktitle={Proceedings of the IEEE/CVF International Conference on Computer Vision},
  pages={3142--3152},
  year={2021}
}

@article{dosovitskiy2020image,
  title={An image is worth 16x16 words: Transformers for image recognition at scale},
  author={Dosovitskiy, Alexey and Beyer, Lucas and Kolesnikov, Alexander and Weissenborn, Dirk and Zhai, Xiaohua and Unterthiner, Thomas and Dehghani, Mostafa and Minderer, Matthias and Heigold, Georg and Gelly, Sylvain and others},
  journal={arXiv preprint arXiv:2010.11929},
  year={2020}
}

@article{li2025recogdrive,
  title={ReCogDrive: A Reinforced Cognitive Framework for End-to-End Autonomous Driving},
  author={Li, Yongkang and Xiong, Kaixin and Guo, Xiangyu and Li, Fang and Yan, Sixu and Xu, Gangwei and Zhou, Lijun and Chen, Long and Sun, Haiyang and Wang, Bing and others},
  journal={arXiv preprint arXiv:2506.08052},
  year={2025}
}

@article{wang2024he,
  title={He-drive: Human-like end-to-end driving with vision language models},
  author={Wang, Junming and Zhang, Xingyu and Xing, Zebin and Gu, Songen and Guo, Xiaoyang and Hu, Yang and Song, Ziying and Zhang, Qian and Long, Xiaoxiao and Yin, Wei},
  journal={arXiv preprint arXiv:2410.05051},
  year={2024}
}

@article{sima2025centaur,
  title={Centaur: Robust end-to-end autonomous driving with test-time training},
  author={Sima, Chonghao and Chitta, Kashyap and Yu, Zhiding and Lan, Shiyi and Luo, Ping and Geiger, Andreas and Li, Hongyang and Alvarez, Jose M},
  journal={arXiv preprint arXiv:2503.11650},
  year={2025}
}

@inproceedings{liao2025diffusiondrive,
  title={Diffusiondrive: Truncated diffusion model for end-to-end autonomous driving},
  author={Liao, Bencheng and Chen, Shaoyu and Yin, Haoran and Jiang, Bo and Wang, Cheng and Yan, Sixu and Zhang, Xinbang and Li, Xiangyu and Zhang, Ying and Zhang, Qian and others},
  booktitle={CVPR},
  pages={12037--12047},
  year={2025}
}

@inproceedings{kendall2019learning,
  title={Learning to drive in a day},
  author={Kendall, Alex and Hawke, Jeffrey and Janz, David and Mazur, Przemyslaw and Reda, Daniele and Allen, John-Mark and Lam, Vinh-Dieu and Bewley, Alex and Shah, Amar},
  booktitle={ICRA},
  pages={8248--8254},
  year={2019},
  organization={IEEE}
}

@inproceedings{jiang2023vad,
  title={Vad: Vectorized scene representation for efficient autonomous driving},
  author={Jiang, Bo and Chen, Shaoyu and Xu, Qing and Liao, Bencheng and Chen, Jiajie and Zhou, Helong and Zhang, Qian and Liu, Wenyu and Huang, Chang and Wang, Xinggang},
  booktitle={Proceedings of the IEEE/CVF International Conference on Computer Vision},
  pages={8340--8350},
  year={2023}
}

@inproceedings{dosovitskiy2017carla,
  title={CARLA: An open urban driving simulator},
  author={Dosovitskiy, Alexey and Ros, German and Codevilla, Felipe and Lopez, Antonio and Koltun, Vladlen},
  booktitle={Conference on robot learning},
  pages={1--16},
  year={2017},
  organization={PMLR}
}

@article{wu2022trajectory,
  title={Trajectory-guided control prediction for end-to-end autonomous driving: A simple yet strong baseline},
  author={Wu, Penghao and Jia, Xiaosong and Chen, Li and Yan, Junchi and Li, Hongyang and Qiao, Yu},
  journal=NIPS,
  volume={35},
  pages={6119--6132},
  year={2022}
}

@article{huang2022cleanrl,
  title={Cleanrl: High-quality single-file implementations of deep reinforcement learning algorithms},
  author={Huang, Shengyi and Dossa, Rousslan Fernand Julien and Ye, Chang and Braga, Jeff and Chakraborty, Dipam and Mehta, Kinal and Ara{\~A}{\v{s}}jo, Jo{\~A}{\c{G}}o GM},
  journal={Journal of Machine Learning Research},
  volume={23},
  number={274},
  pages={1--18},
  year={2022}
}

@article{zhou2024hugsim,
  title={Hugsim: A real-time, photo-realistic and closed-loop simulator for autonomous driving},
  author={Zhou, Hongyu and Lin, Longzhong and Wang, Jiabao and Lu, Yichong and Bai, Dongfeng and Liu, Bingbing and Wang, Yue and Geiger, Andreas and Liao, Yiyi},
  journal={arXiv preprint arXiv:2412.01718},
  year={2024}
}

@article{kerbl20233d,
  title={3D Gaussian splatting for real-time radiance field rendering.},
  author={Kerbl, Bernhard and Kopanas, Georgios and Leimk{\"u}hler, Thomas and Drettakis, George},
  journal={ACM Trans. Graph.},
  volume={42},
  number={4},
  pages={139--1},
  year={2023}
}

@article{liao2022kitti,
  title={Kitti-360: A novel dataset and benchmarks for urban scene understanding in 2d and 3d},
  author={Liao, Yiyi and Xie, Jun and Geiger, Andreas},
  journal=PAMI,
  volume={45},
  number={3},
  pages={3292--3310},
  year={2022},
  publisher={IEEE}
}

@inproceedings{sun2020scalability,
  title={Scalability in perception for autonomous driving: Waymo open dataset},
  author={Sun, Pei and Kretzschmar, Henrik and Dotiwalla, Xerxes and Chouard, Aurelien and Patnaik, Vijaysai and Tsui, Paul and Guo, James and Zhou, Yin and Chai, Yuning and Caine, Benjamin and others},
  booktitle=CVPR,
  pages={2446--2454},
  year={2020}
}

@inproceedings{caesar2020nuscenes,
  title={nuscenes: A multimodal dataset for autonomous driving},
  author={Caesar, Holger and Bankiti, Varun and Lang, Alex H and Vora, Sourabh and Liong, Venice Erin and Xu, Qiang and Krishnan, Anush and Pan, Yu and Baldan, Giancarlo and Beijbom, Oscar},
  booktitle=CVPR,
  pages={11621--11631},
  year={2020}
}

@inproceedings{xiao2021pandaset,
  title={Pandaset: Advanced sensor suite dataset for autonomous driving},
  author={Xiao, Pengchuan and Shao, Zhenlei and Hao, Steven and Zhang, Zishuo and Chai, Xiaolin and Jiao, Judy and Li, Zesong and Wu, Jian and Sun, Kai and Jiang, Kun and others},
  booktitle={ITSC},
  pages={3095--3101},
  year={2021},
  organization={IEEE}
}

@article{shao2024deepseekmath,
  title={Deepseekmath: Pushing the limits of mathematical reasoning in open language models},
  author={Shao, Zhihong and Wang, Peiyi and Zhu, Qihao and Xu, Runxin and Song, Junxiao and Bi, Xiao and Zhang, Haowei and Zhang, Mingchuan and Li, YK and Wu, Yang and others},
  journal={arXiv preprint arXiv:2402.03300},
  year={2024}
}

@inproceedings{codevilla2018end,
  title={End-to-end driving via conditional imitation learning},
  author={Codevilla, Felipe and M{\"u}ller, Matthias and L{\'o}pez, Antonio and Koltun, Vladlen and Dosovitskiy, Alexey},
  booktitle={ICRA},
  pages={4693--4700},
  year={2018},
  organization={IEEE}
}

@article{nehme2023safe,
  title={Safe navigation: Training autonomous vehicles using deep reinforcement learning in carla},
  author={Nehme, Ghadi and Deo, Tejas Y},
  journal={arXiv preprint arXiv:2311.10735},
  year={2023}
}

@article{delavari2025comprehensive,
  title={A Comprehensive Review of Reinforcement Learning for Autonomous Driving in the CARLA Simulator},
  author={Delavari, Elahe and Khanzada, Feeza Khan and Kwon, Jaerock},
  journal={arXiv preprint arXiv:2509.08221},
  year={2025}
}

@article{chen2024end,
  title={End-to-end autonomous driving: Challenges and frontiers},
  author={Chen, Li and Wu, Penghao and Chitta, Kashyap and Jaeger, Bernhard and Geiger, Andreas and Li, Hongyang},
  journal={IEEE Transactions on Pattern Analysis and Machine Intelligence},
  year={2024},
  publisher={IEEE}
}

@inproceedings{dauner2024navsim,
	title = {NAVSIM: Data-Driven Non-Reactive Autonomous Vehicle Simulation and Benchmarking},
	author = {Daniel Dauner and Marcel Hallgarten and Tianyu Li and Xinshuo Weng and Zhiyu Huang and Zetong Yang and Hongyang Li and Igor Gilitschenski and Boris Ivanovic and Marco Pavone and Andreas Geiger and Kashyap Chitta},
	booktitle = NIPS,
	year = {2024},
}

@article{li2024hydra,
  title={Hydra-MDP: End-to-end Multimodal Planning with Multi-target Hydra-Distillation},
  author={Li, Zhenxin and Li, Kailin and Wang, Shihao and Lan, Shiyi and Yu, Zhiding and Ji, Yishen and Li, Zhiqi and Zhu, Ziyue and Kautz, Jan and Wu, Zuxuan and others},
  journal={arXiv preprint arXiv:2406.06978},
  year={2024}
}

@article{wang2025enhancing,
  title={Enhancing autonomous driving safety with collision scenario integration},
  author={Wang, Zi and Lan, Shiyi and Sun, Xinglong and Chang, Nadine and Li, Zhenxin and Yu, Zhiding and Alvarez, Jose M},
  journal={arXiv preprint arXiv:2503.03957},
  year={2025}
}

@article{booher2024cimrl,
  title={Cimrl: Combining imitation and reinforcement learning for safe autonomous driving},
  author={Booher, Jonathan and Rohanimanesh, Khashayar and Xu, Junhong and Isenbaev, Vladislav and Balakrishna, Ashwin and Gupta, Ishan and Liu, Wei and Petiushko, Aleksandr},
  journal={arXiv preprint arXiv:2406.08878},
  year={2024}
}
